\documentclass[10pt,journal,compsoc]{IEEEtran}

\ifCLASSOPTIONcompsoc
  \usepackage[nocompress]{cite}
\else
  \usepackage{cite}
\fi

\ifCLASSINFOpdf

\else
 
\fi

\usepackage{times}

\usepackage{soul}
\usepackage{url}
\usepackage[hidelinks]{hyperref}
\usepackage[utf8]{inputenc}
\usepackage[small]{caption}
\usepackage{graphicx}
\usepackage{amsmath}
\usepackage{booktabs}
\usepackage{xcolor}

\usepackage{amsmath}
\usepackage{xcolor,soul,framed} 
\usepackage{xspace}
\usepackage{array}
\usepackage{eqparbox}
\usepackage{url}
\usepackage{longtable}
\usepackage{lipsum}
\usepackage{blindtext}
\usepackage{makecell}
\usepackage{mathtools}
\usepackage{commath}
\usepackage{multirow}
\usepackage{tabularx}
\usepackage[normalem]{ulem}
\usepackage{xspace}
\usepackage{algorithm}
\usepackage{algpseudocode}
\usepackage{amsfonts}
\usepackage{amssymb}
\usepackage{pifont}
\newcommand{\cmark}{\ding{51}}%
\newcommand{\xmark}{\ding{55}}%

\begin{document}
%

\title{View-Invariant Gait Recognition with Attentive Recurrent Learning of Partial Representations}

\author{Alireza Sepas-Moghaddam \IEEEmembership{Member, IEEE}  and Ali Etemad, \IEEEmembership{Senior Member, IEEE}

\thanks{Alireza Sepas-Moghaddamand and Ali Etemad are with the Department of Electrical and Computer Engineering \& Ingenuity Labs Research Institute, Queen's University, Kingston, ON, K7L 3N6 Canada (e-mail: alireza.sepasmoghaddam@queensu.ca). The authors would like to thank the BMO Bank of Montreal and Mitacs for funding this research. }}

\IEEEtitleabstractindextext{%

\markboth{IEEE Transactions on Biometrics, Behavior, and Identity Science}%
{Shell \MakeLowercase{\textit{et al.}}: Bare Demo of IEEEtran.cls for Biometrics Council Journals}

\begin{abstract}
Gait recognition refers to the identification of individuals based on features acquired from their body movement during walking. Despite the recent advances in gait recognition with deep learning, variations in data acquisition and appearance, namely camera angles, subject pose, occlusions, and clothing, are challenging factors that need to be considered for achieving accurate gait recognition systems. In this paper, we propose a network that first learns to extract gait convolutional energy maps (GCEM) from frame-level convolutional features. It then adopts a bidirectional recurrent neural network to learn from split bins of the GCEM, thus exploiting the relations between learned partial spatiotemporal representations. We then use an attention mechanism to selectively focus on important recurrently learned partial representations as identity information in different scenarios may lie in different GCEM bins. Our proposed model has been extensively tested on two large-scale CASIA-B and OU-MVLP gait datasets using four different test protocols and has been compared to a number of state-of-the-art and baseline solutions. Additionally, a comprehensive experiment has been performed to study the robustness of our model in the presence of six different synthesized occlusions. The experimental results show the superiority of our proposed method, outperforming the state-of-the-art, especially in scenarios where different clothing and carrying conditions are encountered. The results also revealed that our model is more robust against different occlusions as compared to the state-of-the-art methods. 
\end{abstract}

}

\maketitle

\IEEEdisplaynontitleabstractindextext

%
\IEEEpeerreviewmaketitle

\section{Introduction}
Gait, defined as the manner in which individuals walk, includes important human information \cite{R1} which has been used in a wide range of application areas including health\cite{pato, clinic, clinic2}, sport\cite{sport1, sport2}, affect analysis \cite{Ali1,Ali2,Ali3}, and biometrics \cite{R4, R2, nambiar}. In contrast to other biometric modalities such as face \cite{face} and iris \cite{iris}, gait can be captured from far distances with no cooperation of individuals and using ordinary or even low-resolution cameras \cite{R2}. Since the introduction of the first gait recognition system in 1997 \cite{R3}, this field has witnessed incredible progress, and is nowadays dominated by deep learning methods \cite{R4}.

The performance of vision-based gait recognition systems are generally affected by variations in \textit{i}) the appearance of users, for instance wearing a coat or a hat, or even carrying a backpack or a handbag; \textit{ii}) the viewpoints from which the gait sequences have been captured as certain parts of the body may be hidden from a subset of the observed views \cite{B6,view}; and \textit{iii}) occlusions, where the person of interest is partially covered by an obstacle \cite{incomp, bar3}. Gait recognition with deep learning has gained momentum in recent years, providing state-of-the-art recognition accuracy \cite{R4}. However, due to the challenging factors stated above, it is hard to learn gait features that are highly robust to occlusions, view-point variations, and appearance changes \cite{B6}. Additionally, unconstrained variations \cite{TanmayThesis} such as changes in lighting \cite{R7} and unconstrained backgrounds \cite{background}, further complicates gait recognition and hinders the performance of the proposed models. 

Gait analysis methods based on partial representations have been widely proven to be effective \cite{partial5,partial6,partial7,partial9,partial10, B6, B1}. These methods are based on splitting gait data into spatial and/or temporal bins, possibly at multiple scales, to obtain partial gait representations. The partial representations are generally concatenated in a single feature vector to perform gait recognition \cite{B6,R9,B1}. However, these representations could be further processed to explore their relations, while preserving positional attributes, such as location, rotation, and scale of the representations. By preserving positional attributes, gait recognition systems can become more robust to view and orientation changes of gait sequences, proving to be instrumental for designing improved view-invariant gait recognition systems \cite{B1}. Additionally, different parts of the body at different instances within the gait sequence may contain varying amounts of person-specific information, which most current solutions based on partial representations fail to exploit. This is due to the fact that partial learned features are often concatenated \cite{part}, assigning similar importance to different parts of the learned embedding. Learning of partial features' importance is more instrumental notably when an object such as a backpack results in a change in appearance.

In this paper, we propose a novel method for learning view-invariant convolutional gait energy maps with an attention-based recurrent model to tackle the above-mentioned problems. In this context, our solution first learns convolutional maps from frame-level gait silhouettes and uses a temporal pooling layer to learn spatiotemporal information by aggregating frame-level representations. The output of this layer, the so called Gait Convolutional Energy Map (GCEM), is then split into distinct bins and fed to a fully-connected layer for dimensionality reduction, thus creating a sequence of partial features. We then exploit a network of bi-directional gated recurrent units (BGRU) \cite{GRU1} to learn from the sequences of partial gait features, thus exploiting the relations between different spatiotemporal parts of the embedding that strongly contribute to gait recognition. We finally use an attention layer to selectively focus on the most important recurrent representations. A softmax layer is used to provide the final set of features for classification based on a cosine proximity loss function.

Our key contributions can be summarized as follows:

\begin{itemize}
    \item For the first time in this context, we propose an attentive recurrent model to learn from gait convolutional energy maps, thus exploiting the relations between different partial representations, both spatially and temporally;
    \item Experiments on large-scale CASIA-B and OU-MVLP datasets indicate that our proposed model performs robustly not only in cross-view settings, but also in different clothing and carrying conditions, setting new state-of-the-art for both datasets. Experiments also reveal that our model is more robust against different synthesized occlusions compared to the other state-of-the-art methods.
\end{itemize}

The rest of the paper is organized as follows. In the next section, we describe the related works and recent research on gait recognition and datasets. Next, we present our proposed model, followed by a description of the experiments performed to evaluate our method. Then, the results are provided along with a comparison to the state-of-the-art and ablation experiments are discussed in detail. Lastly, the summary is presented.

\section{Related Work}
This section provides an overview of deep learning methods and datasets for gait recognition. 

\begin{table*}
  \centering
  \setlength\tabcolsep{8pt}
    \caption{Overview of state-of-the-art deep gait recognition methods.}
    \begin{tabular}{l|l|l|l|l|l}
    \hline
    \textbf {Method}& \textbf{Year}& \textbf{Input}& \textbf{Representation} & \textbf{Network Architecture} & \textbf{Temporal Template} \\
    \hline\hline
    GEINet~\cite{B8} & 2016& Silhouette & Global &   2 Convolution +  2 Pooling+ 2 FC& GEI \\
    \hline
    {\vtop{\hbox{\strut Ensemble}\hbox{\strut CNNs~\cite{B6}}}} & 2017 & Silhouette & {\vtop{\hbox{\strut Global}\hbox{\strut Partial}}}   &  {\vtop{\hbox{\strut 3 Convolution +  2 Pooling + 2 FC}\hbox{\strut 2 Convolution +  2 Pooling + 1 FC}}}& {\vtop{\hbox{\strut GEI}\hbox{\strut CGIs}}} \\
    \hline
    DiGGAN~\cite{B9} & 2018& Silhouette & Global  &  Auto-encoder & GEI 
      \\
    \hline
    MGANs ~\cite{B7} & 2019& Silhouette & Global &  4 Convolution +  1 Pooling + 3 FC& PEI  \\
    \hline
    EV-Gait~\cite{B2} & 2019& Silhouette & Global &   3 Convolution +  3 Pooling + 3 Capsule & GEI  \\
    \hline
    Gait-Joint ~\cite{B10} & 2019 & Silhouette & Global  &  12 Convolution +  5 Pooling + 3 FC & GEI  \\
    \hline
    Gait-Set ~\cite{B1} & 2019& Silhouette & Partial &   6 Convolution +  2 Pooling + 1 FC& Set-pooling Maps \\
    \hline
    GaitNet-1 ~\cite{B3}& 2019&  Silhouette; Skeleton & Global &   Auto-encoder + LSTM& N/A \\
    \hline
    GaitNet-2~\cite{B4} & 2020& Silhouette; Skeleton & Global &   Auto-encoder + LSTM & N/A \\
    \hline
    \end{tabular}%
  \label{tab1}%
\end{table*}%

\begin{table*}
  \centering
  \setlength\tabcolsep{3pt}
    \caption{Overview of the gait datasets with different characteristics.}
    \begin{tabular}{l|l|l|l|l|l|l}
    \hline
    \textbf {Dataset}& \textbf{Year}& \textbf{Data Type} & \textbf{\# of Subjects}& \textbf{\# of Sequences} & \textbf{\# of Views} & \textbf{Variations} \\
    \hline\hline
     CMU MoBo \cite{DB-MOBO}& 2001 & RGB; Silhouette & 25 & 600 & 6 & 3 Walking Speeds; Carrying a Ball\\
    \hline
    SOTON Large \cite{DB-SOTON} & 2002 & RGB; Silhouette & 115 & 2,128 & 2 & Indoor \& Outdoor; Treadmill\\
    \hline
    Georgia Tech \cite{DB-Georgia} & 2003& 3D RGB & 15 & 540 & 1 & 4 Walking Speeds\\
    \hline
    CASIA A \cite{DB-CASIAA}& 2003 & RGB & 20 & 240 & 3 & Normal Walking\\
    \hline
    USF HumanID \cite{DB-HumanID} & 2005 & RGB & 122 & 1,870 & 2 & Outdoor Walking; Carrying a Briefcase; Time Interval\\

    \hline
    CASIA B \cite{CASIA} & 2006 & RGB; Silhouette & 124 & 13,680 & 11 & Normal Walking; Carrying a Bag; Wearing a Coat\\
    
    \hline
    CASIA C \cite{DB-CASIAC} & 2006 & Infrared; Silhouette & 153 & 1,530 & 1 & 3 Walking Speeds; Carrying a Bag \\
    \hline
    OU-ISIR LP \cite{DB-OU}  & 2012 & Silhouette & 4,007 & 31,368 & 4 & Normal Walking\\
    \hline
    TUM GAID \cite{DB-TUM}  & 2012 & RGB; Depth; Audio & 305  & 3,737 & 1 & Normal Walking\\
    \hline
    KY IR Shadow \cite{DB-Shadow} & 2014 & Shadow Silhouette & 54 & 324 & 1 & Normal Walking; Carrying a Bag; Changing the Clothes\\
    \hline
    OU-MVLP \cite{MVLP}  & 2017 & Silhouette & 10,307  & 259,013 & 14 & Normal Walking\\
    \hline
    
    \end{tabular}%
  \label{tabDB}%
\end{table*}%

\subsection{Deep Gait Recognition Methods}
Gait recognition methods can be categorized into model-based \cite{Model} and appearance-based \cite{B6} categories. The methods in the first category typically model the body structure during walking, for example by extracting the human body skeleton. Appearance-based methods, on the other hand, may not explicitly consider the body structure and are generally more robust to situations where the human body structure may not be accurately modeled under uncontrolled conditions, e.g., in the presence of appearance variations. Gait recognition methods that lie within the former category first extract gait information, in the form of gait silhouettes, from different video frames. They then model temporal template representations, for instance the average silhouette over a gait cycle, also known as Gait Energy Image (GEI), thus aggregating temporal walking information. Finally, in these methods, a classifier can be used to perform identity recognition. Our gait recognition method proposed in this paper belongs to the second category.

Table \ref{tab1} presents an overview of the main characteristics of deep learning methods for gait recognition, sorted according to their release date, highlighting their input type, exploited representation, network architectures, and utilized temporal templates. Deep gait recognition methods accept as their inputs different types of gait information, such as gait silhouette or body joint/skeleton information, to be processed by the network. These methods can learn different representations, including global representations that deal with the appearance and/or motion of the body as a whole, or alternatively as partial representations, which split the learned representations of the body into different bins for further processing. These temporal templates can be obtained either in the initial layer of the deep network in the form of Gait Energy Images (GEI) \cite{B6,R7}, Period Energy Images (PEI) \cite{B7}, Chrono Gait Images (CGI) \cite{GEI1}, and others, or in an intermediate layers of the network, for instance in the form of set-pooling maps \cite{B1}. 

The current state-of-the-art on gait recognition is dominated by deep learning methods. In one of the early deep learning works in the area, GEINet \cite{B8} was proposed in 2016, consisting of a set of layers, to perform gait recognition. GEINet was fed by GEIs and subsequently two sequential triplets of convolutional, pooling, normalization along with two fully connected layers were used to produce the final embeddings. A softmax classifier was finally used to perform gait recognition. In \cite{B6}, three different deep CNN architectures were proposed, with variations in both depth and architecture, whose inputs were GEIs. An ensemble of these networks were finally proposed in order to obtain discriminative gait features for performing classification. Discriminant gait generative adversarial network (DiGGAN) was proposed in \cite{B9}, exploiting a generator and a discriminator to extract robust view-invariant features while preserving the identity information. In this context, DiGGAN aimed to transfer a GEI image captured at an arbitrary view to a target view. As the arbitrary and target view images share the same identity, an auto-encoder can be applied in order to disentangle the view and the identity information. This can then be used to transfer an image into a new latent space, while preserving identity information. In \cite{B7}, a multi-task generative adversarial network (MGANs) was proposed using a new multi-channel gait temporal template, PEI, capable of learning view-invariant gait representations. MGANs includes: \textit{i}) an encoder that produces gait view-specific features in a latent space; \textit{ii}) a view classifier that predicts the angle for the view-specific features; \textit{iii}) a transformer to map the view-specific features from one view to another; \textit{iv}) a generator to generate the view-specific gait images; and \textit{v}) a discriminator that decides if the generated image belongs to certain distributions. 

Event-based gait Recognition (EV-Gait) \cite{B2} exploited motion consistency available in a dynamic vision sensor using a CNN. EV-Gait includes convolutional layers with residual blocks for performing feature extraction, and utilizes fully-connected layers with softmax classifier to distinguish between different identities. Gait-Joint \cite{B10} was proposed based on two CNNs for segmentation and classification. Gait-Joint first extracts gait silhouettes using a fully convolutional network and uses a deep network to combine the silhouette sequences for the recognition purpose. Finally, segmentation and recognition networks were jointly learned to extract more discriminate features for gait recognition. GaitSet \cite{B1} was proposed as an end-to-end deep model, extracting set-level features. GaitSet first used a CNN to extract gait features from gait silhouettes to be then attentively combined to form set-pooling maps. Finally, a so called horizontal pyramid mapping scheme, adopted from \cite{R9}, was applied for learning partial representations at multiple scales. The partial features were finally concatenated to feed the classifier. A novel auto-encoder network was proposed in \cite{B3} to extract discriminative features from different views to be then fed to a multi-layer LSTM network for integrating pose features over time. Finally, the same authors extended their initial idea by modifying the training scheme, notably by changing the loss function in \cite{B4}.

\subsection{Gait Analysis Based on Partial Representations}
Gait analysis methods with partial representations generally split the gait data into spatial and/or temporal bins \cite{partial5,partial6,partial7,partial9, partial10, B1}. The obtained partial gait representations are then further possessed and fused to perform gait analysis tasks. In \cite{partial5}, gait silhouettes were divided into 7 parts. First the silhouettes were divided into 4 horizontal bins, and then the horizontal bins, except the top bin, were further divided into two parts with the same size in the vertical direction. Moment-based features were finally extracted from each part and were concatenated as the input to the classifier. The method proposed in \cite{partial6} aimed to perform gait-based gender classification by segmenting GEIs into head and hair, chest, back, waist, and buttocks. A weighting scheme was then used to assign higher weights to the most important parts and a Support Vector Machine (SVM) was used for gender classification. In \cite{partial7}, gait silhouettes were divided into eight parts including head, torso, 2 thighs, 2 arms, and 2 calves/feet. Next, each part was fitted by an ellipse model to extract some structural parameters including the centroid's \textit{x} and \textit{y} coordinates, the orientation of the axes, and the aspect ratio of the ellipse. The features were finally concatenated to create the final feature vector to feed the classifier. The method proposed in \cite{partial9} divided the GEI into five parts including head, chest, back, waist and buttocks, and legs. Linear Discriminant Analysis was then used to extract feature vectors from these parts. In \cite{partial10}, gait GEIs were split into four segments. This was adaptively done using Genetic Algorithm to optimize the split boundaries for dividing these segments as well as a binary weight per each segment to decide if the respective segment should be included in the training or not. Finally, in \cite{B1}, the approach entitled GaitSet split the set-pooling maps into multi-scale partial representations to be then used as input to a softmax classifier.

Our proposed method in this paper is also based on deep partial representations. The method learns, for the first time, the relations between partial representations using an RNN and then learns the importance of each representation using an attention mechanism. None of the above mentioned methods have used a similar approach to improve the discrimination of the partial representations; instead they concatenate these representations to feed the classifier. Additionally, our method segments the GCEMs into 16 horizontal partial bins, while other deep methods including \cite{R9} and \cite{B1} split the temporal templates into multi-scale partial representations that involves higher computational complexity and larger feature maps.

\subsection{Gait Datasets}
Different gait datasets have been developed to evaluate gait recognition systems. These datasets have been collected to cover different variations, both in terms of the appearance of the subjects and acquisition viewpoints. They should also be large enough to deliver reliable results, notably when deep learning solutions are used. Table \ref{tabDB} overviews the main characteristics of existing notable gait datasets including the type of data, the number of subjects, the total number of sequences, the number of acquisition viewpoints, and the variations in the gait sequences. These  datasets  are  sorted  in  the  table by the order of release date, thus showing the chronological evolution of the gait datasets.

As can be observed in Table \ref{tabDB}, CASIA B \cite{CASIA} and OU-MVLP \cite{MVLP} have been obtained with the most number of acquisition viewpoints. As our proposed method has been designed to learn view-invariant gait features, we have used these two large-scale datasets in our experiments. Additionally, CASIA B \cite{CASIA} dataset includes different clothing and carrying variations that facilitate gait recognition experiments against appearance changes. Due to the large number of variations available in these dataset, they can be suitable to evaluate the recognition performance with high reliability.

\section{Proposed Method}
This section presents the proposed method for view-invariant gait recognition, including the model intuition, architecture, and walkthroughs.

\subsection{Problem Definition}\label{sec:problem_definition}

Consider a dataset $\mathcal{D}$ of gait silhouette sequences where each sequence $S_{i, j}$ is recorded from identity $x_i \in \{x_1,x_2,\dots, x_N\}$ and recording view angle $\theta_j \in \{\theta_{1}, \theta_{2}, \dots, \theta_{V}\}$. The gait sequences available in dataset $\mathcal{D}$ can be divided into $Training$, $Gallery$, or $Probe$ sets. The task of gait recognition for a probe gait sequence $Probe_{i, j}$ can then be defined as:
\begin{equation} \label{eq:task}
    \hat{x} = arg \max\limits_{x_i} Pr(x_i|Probe_{i, j}, Gallery),
\end{equation}
where $Pr(x_i|Probe_{i, j}, Gallery)$ is the probability of sequence $Probe_{i, j}$ belonging to identity $x_i$ in the $Gallery$ set. In view-invariant gait recognition, all the combinations of views are considered during testing, except identical angles between $Gallery$ and $Probe$ sets.

\subsection{Model Intuition}
The intuitive approach for gait recognition is based on extracting features from the whole body available in either gait silhouettes or temporal templates. These methods aim to capture the most salient cues of the global appearance of the body in order to represent identities of different individuals. However, global representations imply that some infrequent details or non-salient information might be ignored, thus making it hard to learn intra-class and inter-class variations.
 
Inspired by the sate-of-the-art person re-identification methods \cite{R9, Partial2, partial3, partial4}, a few gait recognition solutions \cite{B6} \cite{B1} have recently taken advantage of deep partial features representations. In this context, they split either gait silhouettes or temporal templates into several bins for further processing. These partial representations are generally concatenated in a single feature vector to perform gait recognition \cite{B6,B1}. However, instead of feature concatenation, the partial representations could be further exploited to learn relations between these parts. It also preserves positional attributes, such as location, rotation, and scale of the representations that can make gait recognition systems more robust against view and orientation changes. Additionally, the importance of each part towards the final recognition performance could be attentively learned by assigning higher weights to the more relevant features while ignoring the spurious dimensions. These issues are the main problems tackled in this paper. 

\subsection{Model Overview}
Given a multi-view gait dataset containing information from $N$ subjects with identities $i = 1,...,N$ and captured from $V$ different views, our goal is to learn a model using two subnetworks. First, we aim to learn partial spatiotemporal features from the gait silhouette sequence, $S_{i, j}$, captured from subject $i$ and view $j$. This can be represented as:
\begin{equation}
\begin{split}
\label{eq:1}
PF_{i,j,b}= \: G_{1}(S_{i,j}): 
\forall_{i}\in(1,...,N); \forall_{j}\in(1,...,V),
\end{split}
\end{equation}
where $PF$ represents the corresponding partial features, obtained by the first subnetwork, $G_1$. The partial features are represented in $B$ bins, indexed by $b: \forall_{b}\in(1,...,B)$. Subnetwork $G_1$ can be broken down to several functions that will be introduced by Equations \ref{eq:4} to \ref{eq:7} in the following subsection.

In the second subnetwork, we aim to exploit the recurrent relations between different segments within the learned spatiotemporal partial features to utilize additional latent identity-related information. This subnetwork can be formulated as:
\begin{equation}
\begin{split}
\label{eq:2}
AF_{i,j}= \: G_{2}(PF_{i,j,b}): \\
\forall_{i}\in(1,...,N);\:
\forall_{j}\in(1,...,V);\:
\forall_{b}\in(1,...,B).
\end{split}
\end{equation}
where $AF$ denotes the corresponding attentive features, output by the second subnetwork $G_2$. Subnetwork $G_2$ can be broken into several functions that will be introduced in Equations \ref{eq:8} to \ref{eq:11}.

\subsection{Model Architecture}
\begin{figure*}[!t]
    \begin{center}
    \includegraphics[width=1\linewidth]{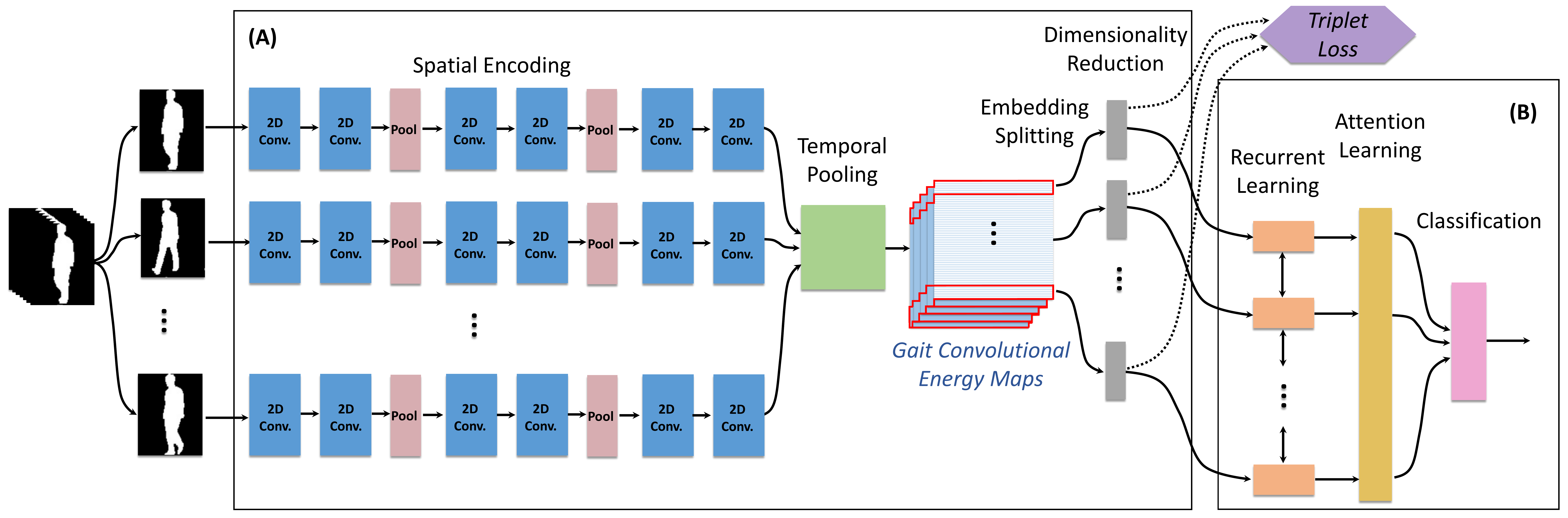} 
    \end{center}
\caption{Architecture of the proposed gait recognition network.}
\label{fig:2}
\end{figure*}

Here, we describe the structure of our deep network for view-invariant gait recognition, as shown in Figure \ref{fig:2}. Our model uses partial representations by splitting the learned gait representations in the form of convolutional energy maps, into separate bins. Then, for the first time, we recurrently learn the split bins to exploit the relation between partial spatiotemporal representations. Additionally, we use an attention mechanism to selectively focus on important recurrently learned parts of the embedding as identity information in different scenarios may lie in different bins. In the following, we describe our model, which consists of a total of fourteen layers.

\subsubsection{Spatial Encoding} First, in a process called \textit{spatial encoding}, our model computes spatial features for each gait silhouette frame $F_{i,j,t}$ captured from subject $i$ at viewing angle $j$ and at time $t$, separately using convolution and pooling layers. In this context, spatial feature $SF_{i,j,t}$ belonging to gait silhouette frame $F_{i,j,t}$ is convolved with $K$ convolution filters. This process can be formulated as:
\begin{equation}
\begin{split}
\label{eq:4}
SF_{i,j,t} = SE(F_{i,j,t},k): \;
\\ \forall_{i}\in(1,...,N); \forall_{j}\in(1,...,V);\\ \forall_{t}\in(1,...,T); \forall_{k}\in(1,...,K),
\end{split}
\end{equation}
where $SE$ is the spatial encoding function, and $N$, $V$, $T$, and $K$ are the number of subjects, views, silhouettes, and filters that have been used for convolution operations respectively. Inspired by \cite{B1} where an eight-layer CNN architecture was used to learn strong spatial representations for gait recognition, we have used eight 2D convolution and pooling layers in the spatial encoding stage, as presented in Table \ref{tabspat}. The output dimension of the last convolution layer for the whole gait silhouette sequence is $T \times 16 \times 16 \times 256$.

\begin{table}
  \centering
  \setlength\tabcolsep{3pt}
  \footnotesize
  \caption{Details of the spatial encoding layers.}
    \begin{tabular}{l|l|l|l|l|l}
    \hline
    \textbf{Layer} & \textbf{\vtop{\hbox{\strut Input}\hbox{\strut Size}}} & \textbf{\vtop{\hbox{\strut No. of}\hbox{\strut Filters}}} & \textbf{\vtop{\hbox{\strut Filter}\hbox{\strut Size}}}& \textbf{Padding} &\textbf{\vtop{\hbox{\strut Output}\hbox{\strut Size}}}\\
    \hline\hline
    Conv 1 & 64$\times$64& 64& 5$\times$5 & 2 & 64$\times$64$\times$64\\
    Conv 2 & 64$\times$64$\times$64& 64& 3$\times$3 & 1 & 64$\times$64$\times$64\\
    Pool 1 & 64$\times$64$\times$64& 64& 2$\times$2 & 1 & 32$\times$32$\times$64\\
    Conv 3 & 32$\times$32$\times$64& 128& 3$\times$3 & 1 & 32$\times$32$\times$128\\
    Conv 4 & 32$\times$32$\times$128& 128& 3$\times$3 & 1 & 32$\times$32$\times$128\\
    Pool 2 & 32$\times$32$\times$128& 128& 2$\times$2 & 1 & 16$\times$16$\times$128\\
    Conv 5 & 16$\times$16$\times$128& 256& 3$\times$3 & 1 & 16$\times$16$\times$256\\
    Conv 6 & 16$\times$16$\times$256& 256& 3$\times$3 & 1 & 16$\times$16$\times$256\\
    \hline
    \end{tabular}%
  \label{tabspat}%
\end{table}%

\subsubsection{Temporal Pooling} 
In general, the order of gait information could provide value for gait recognition \cite{CVPRNew}. However, in methods such as ours in which gait silhouettes are averaged to obtain a temporal template, preserving the order of body silhouettes is not advantageous.
In different architectures, temporal templates can generally be obtained either in the initial layer or in an intermediate layer of the model. Our proposed solution in this paper exploits the latter strategy. Here, in order to represent human motion in a single map while preserving temporal information, gait convolutional energy maps (GCEM) have been considered by aggregating convolutional features over time dimension $t$ using Equation \ref{eq:5}.
\begin{equation}
\begin{split}
\label{eq:5}
GCEM_{i,j}=mean^{t}(SF_{i,j,t}):\\
 \forall_{i}\in(1,...,N);\:  \forall_{j}\in(1,...,V);\:  \forall_{t}\in(1,...,T).
\end{split}
\end{equation}

This temporal pooling operation makes the proposed method independent of the number of frames. Therefore the number of frames per each gait sequence may be different. It should be noted that in order to utilize all the available gait data, we aggregate the convolutional features over all the sequence frames. Accordingly, the output dimension of this layer for the whole gait silhouette sequence is $16 \times 16 \times 256$. 

\subsubsection{Embedding Splitting} Partial representations offer some advantages for gait recognition, including reduced sensitivity to missing key body parts and appearance variations \cite{B6}\cite{B1}, as well as ignoring of infrequent details and non-salient information \cite{R9}. In order to exploit these advantages, the GCEM is horizontally split using function $SP$, into $B$ bins (as shown in Figure \ref{fig:2}):
\begin{equation}
\begin{split}
\label{eq:6}
Bin_{i,j,b} = SP(GCEM_{i,j}):\\
\forall_{i}\in(1,...,N);\:
\forall_{j}\in(1,...,V);\:
\forall_{b}\in(1,...,B).
\end{split}
\end{equation}

The output dimension of this layer is $B \times \frac{16}{B} \times 16 \times 256$. For the purpose of this paper, $B = 16$ achieved the best performance, which will be discussed in Section 5.4 (ablation experiments).

\subsubsection{Dimensionality Reduction} The extracted spatiotemporal bins have a high dimensionality, which in turn require a relatively deep model for further processing (for example by the recurrent learning layer). To address this problem, each extracted GCEM bin is independently used as an input to a fully connected (FC) layer with \textit{ReLu} activation functions, learning non-linear combination of the input bin features in a higher-level lower-dimensional latent space. This process outputs Partial Features (PF) from FC layer and can be formulated as:
\begin{equation}
\begin{split}
\label{eq:7}
{PF}_{i,j,b} = DR(Bin_{i,j,b}):\\
\forall_{i}\in(1,...,N);\:
\forall_{j}\in(1,...,V);\:
\forall_{b}\in(1,...,B).
\end{split}
\end{equation}
where $DR$ denotes the dimensionality reduction function to produce features, $PF$, obtained by the fully connected layer. The output size of the FC layer is set to $256$ per each of the $16$ bins, resulting in a $16 \times 256$ output.

\subsubsection{Recurrent Learning} 
To take advantage of the relations between appearance changes corresponding to the partial spatiotemporal GCEM features that are likely to contain person-specific information, we feed the features learned in the previous step to a network of bi-directional gated recurrent units (BGRU) \cite{GRU1} to recurrently learn the relations between partial spatiotemporal representations. The BGRU network has been proven to be an efficient model for sequence learning \cite{GRU2}. This network captures both forward and backward relationships within a partial representation space and can be modeled as two independent conventional GRU networks, respectively analyzing the input sequence of partial representations in forward and backward directions. This layer can be formulated as:
\begin{equation}
\begin{split}
\label{eq:8}
H_{i,j,b} = GRU({PF}_{i,j,b}):\\
\forall_{i}\in(1,...,N);\:
\forall_{j}\in(1,...,V);\:
\forall_{b}\in(1,...,B);
\end{split}
\end{equation}
where the $GRU$ function independently extracts two forward and backward hidden feature vectors. These feature vectors, each including $B$ partial representations, are finally concatenated to form the bi-directional feature vector.

A GRU network is composed of \textit{GRU cells} with a shared \textit{hidden state} memory to keep the dependencies over the input sequence. This shared memory is controlled by two \textit{update} and \textit{reset} gates in the form of two vectors. These gates allow the network to update the shared memory with respect to the new information, and to decide when to reset the previous state, respectively. The updated hidden state corresponding to the input information then produces the output for the GRU cell. The output of each GRU cell is used as input to the next cell, thus creating a network of GRU cells.
The function $GRU$ can be broken into several functions that will be introduced in Equations \ref{eq:GRU1}-\ref{eq:Out}.

For partial feature $PF_{i,j,b}$, where $i$ is the subject's ID, $j$ is the viewing angle, and $b$ is the bin's number, the update gate $Z$ is computed based on $PF_{i,j,b}$ and the previous hidden state, $H_{i,j, b-1}$, to learn to learn how much the cell updates the hidden state: 
\begin{equation}
\label{eq:GRU1}
{Z}_{i,j,b}=\sigma({W_Z}[PF_{i,j,b}+H_{i,j,b-1}]+{b_{Z}}),
\end{equation}
where ${W_Z}$ is the update gate weight and $b_{Z}$ is the update gate bias. $\sigma$ is a sigmoid activation function, defined as $\sigma(x)={(1+{e}^{-x})}^{-1}$, that bounds the output to the [0,1] range. It is worth noting that the input hidden state to the first GRU cell is set to zero.

The reset gate, $R_{i,j,b}$, for partial feature $PF_{i,j,b}$, can similarly be computed based on $PF_{i,j,b}$ and the previous hidden state, $H_{i,j,b-1}$, to 
remove the information not relevant to the prediction from the shared memory according to Equation \ref{eq:GRU2}:
\begin{equation}
\label{eq:GRU2}
{R}_{i,j,b}=\sigma({W_R}[PF_{i,j,b}+H_{i,j,b-1}]+{b_{R}}),
\end{equation}
where ${W_R}$ is the reset gate weight and $b_{R}$ is the reset gate bias. 

The update gate's parameters (${W_Z}$ and ${b_Z}$), and the reset gate's parameters (${W_R}$ and ${b_R}$) have been independently calculated. As a result, these gates can be computed in parallel or in any chosen order.

Next, the vector of the hidden state candidates, ${\tilde{H}_{i,j,b}}$, which will be added later to the hidden state, is formed as:
\begin{equation}
\label{eq:GRU3}
\begin{split}
\tilde{H}_{i,j,b}=\tanh({W_{\tilde{H}}}[PF_{i,j,b} + (R_{i,j,b} \odot H_{i,j,b-1})]+{b_{\tilde{H}}}),
\end{split}
\end{equation}
where $W_{\tilde{H}}$ and $b_{\tilde{C}}$ are respectively the weight and bias for the vector of the hidden state candidates, $\odot$ denotes element-wise multiplication, and the tangent hyperbolic activation function can be formulated as $\tanh(x)=2\sigma(2x)-1$.
In Equation \ref{eq:GRU3}, the past hidden memory information has no contribution toward creating the vector of the hidden state candidates if the reset gate is set to zero, whereas equal contribution is maintained if set to 1. Values between 0 and 1 describe the portion of the information that contribute to create the vector of the hidden state candidates \cite{GRU1}.

Finally, the hidden state can be formulated as:
\begin{equation}
\label{eq:Out}
{H}_{i,j,b}= (1-{Z}_{i,j,b}) \times H_{i,j,b-1} + {Z}_{i,j,b} \times \tilde{H}_{i,j,b},
\end{equation}
where ${H}_{i,j,b}$ is the cell output for the sequence component $PF_{i,j,b}$ which also feeds the the next GRU cell. The number of cells in a GRU network equals the number of bins, i.e., 16. We also set the hidden layer size of the GRU network to 128, so the output size of the BGRU layer is $16 \times 256$.

\subsubsection{Attention Learning} 
Different partial GCEM representations may not equally contribute to the gait recognition performance. For example, a GCEM bin representing convolutional maps of a backpack may negatively affect the performance, while the lower-body bins or waving arms may contribute more. To this end, an attention layer is employed to learn from the extracted BGRU features, thus selectively focusing on the most important features \cite{ATT1, ICASSP}. The attention layer multiplies the BGRU outputs by the corresponding trainable weights \cite{ATT2}. Equations \ref{eq:9} and \ref{eq:10} learn the scores $a_{i,j,b}$ to measure the attention level for $H_{i,j,b}$: 
\begin{equation}
\begin{split}
\label{eq:9}
u_{i,j,b}=tanh(W_{H}H_{i,j,b}+b_{H}):\\
\forall_{i}\in(1,...,N);\:
\forall_{j}\in(1,...,V);\:
\forall_{b}\in(1,...,B);
\end{split}
\end{equation}
\begin{equation}
\begin{split}
\label{eq:10}
a_{i,j,b}=\frac{e^{u_{i,j,b}}}{\sum_{k=1}^{B}{e^{u_{i,j,k}}}}:\\
\forall_{i}\in(1,...,N);\:
\forall_{j}\in(1,...,V);\:
\forall_{b}\in(1,...,B);
\end{split}
\end{equation}
where $W_{H}$ and $b_{H}$ are the trainable weights and biases for the hidden state, respectively.

Finally, $H_{i,j,b}$ is multiplied by the corresponding $a_{i,j,b}$ to calculate the weighted recurrent features. The final attentive feature, $AF$, with the size of $4096$ ($16 \times 256$) for subject $i$ and view $j$, is formed by concatenating all the weighted recurrent features: 
\begin{equation}
\begin{split}
\label{eq:11}
AF_{i,j} = \Vert_{b=1}^{B}({a_{i,j,b}} \times H_{i,j,b});\\
\forall_{i}\in(1,...,N);\:
\forall_{j}\in(1,...,V);\:
\forall_{b}\in(1,...,B);
\end{split}
\end{equation}
where $\Vert$ denotes the concatenation operation over the weighted recurrent features. This means that all the weighted recurrent features for a specific subject at a certain angle are concatenated in a single feature vector.

\subsubsection{Classification} 
The final embedding for subject $i$ and view $j$, $AF_{i,j}$, is then used as input to a dense layer with softmax activation function to perform the classification.

\subsection{Training Loss}
In order to better tune the hyper-parameters of the network and reduce the computational complexity, we trained our model in two separate stages. First, we trained the first subnetwork, illustrated in Figure \ref{fig:2}(A), using triplet loss \cite{triplet}. In this context, three parallel instances of the first subnetwork with shared parameters are used. These instances are fed with three types of inputs: \textit{i)} an anchor sample, \textit{ii)} a positive sample which has the same identity as the anchor, and \textit{iii)} a negative sample which has a different identity as the anchor. A loss value for each triplet is then defined as the $L2$ distance by at least a margin of $M$ between the concatenated feature vectors ${PF}$ (Equation \ref{eq:7}) of the anchor and the positive as well as the negative samples using: 
\begin{equation}
\begin{split}
\label{eq:tiplet}
L_{triplet}=\frac {1}{2M} Max( M-L2_{anch, pos}+L2_{anch, neg},0)\
\end{split}
\end{equation}

This triplet loss function ensures that the dissimilarity between a feature vector belonging to a subject and another vector for the same subject is lower than that between a feature vector belonging to a subject and another vector for a different subject, by at least a margin $M$ \cite{triplet}. In our experiments, we set $M = 0.2$ as recommended in \cite{TS}. In order to sample the triplets for training, we used online triplet mining \cite{triplet} for each batch of the inputs. Given a batch of $B$ concatenated feature vectors $PF$ corresponding to $B$ gait sequences, $B^{3}$ triplets can be computed where most of them are invalid as they do not exactly include one anchor, one positive, and one negative sample. The valid triplets can be divided into three categories: \textit{i)} easy triplets that have a loss value of 0 meaning that the negative sample has sufficient distance to the anchor sample; \textit{ii)} semi-hard triplets that have a positive loss value, but the positive sample is closer to the anchor sample than the negative sample; \textit{iii)} hard triplets that have a positive loss value and the negative sample is closer to the anchor sample than the positive sample. In our experiments, we have used the Batch All (BA+) strategy \cite{TS} to select all the valid triplets, followed by averaging the loss values on the semi-hard and hard triplets.

Once the first subnetwork is trained, we extract $PF$ from the fully connected layer to be used as inputs to the second subnetwork of the model, illustrated in Figure \ref{fig:2}(B). For the second subnetwork, we used a \textit{cosine proximity} \cite{cosine} loss function with \textit{Adam} optimizer for training. The convergence of the second subnetwork was faster than the first one, as the second subnetwork includes a considerably smaller number of learnable parameters.

\subsection{Testing}
Once the network is fully trained with the training data, gallery and probe images are used as inputs to our trained model during testing, thus extracting embeddings as the output of the attention layer. We then used a softmax classifier with \textit{cosine proximity} loss function to compare the probe features with the gallery ones in order to identify the most similar gait sequence.

\begin{figure*}[!t]
    \begin{center}
    \includegraphics[width=0.9\linewidth]{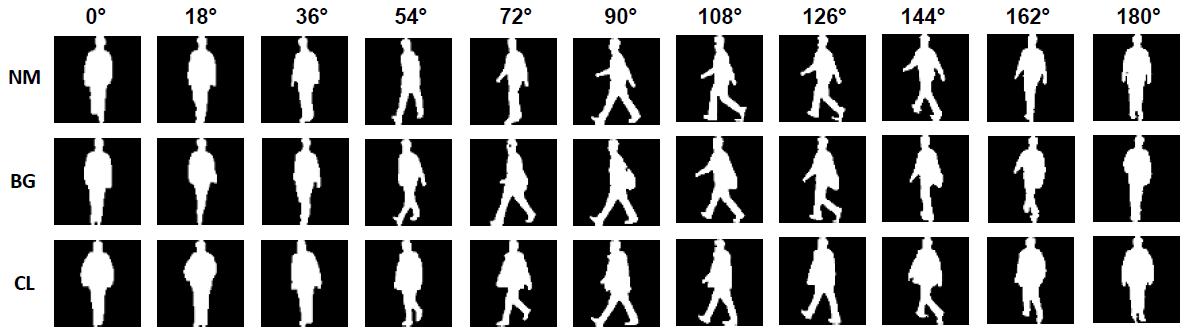} 
    \end{center}
\caption{Example cropped gait silhouettes from one subject in CASIA-B dataset \cite{CASIA}, captured in 11 evenly spaced angles over $0^\circ$ to $180^\circ$ ($0^\circ$, $18^\circ$, $36^\circ$, \dots, $180^\circ$) for three different walking conditions.}
\label{fig:casia}
\end{figure*}

\begin{figure*}[!t]
    \begin{center}
    \includegraphics[width=1\linewidth]{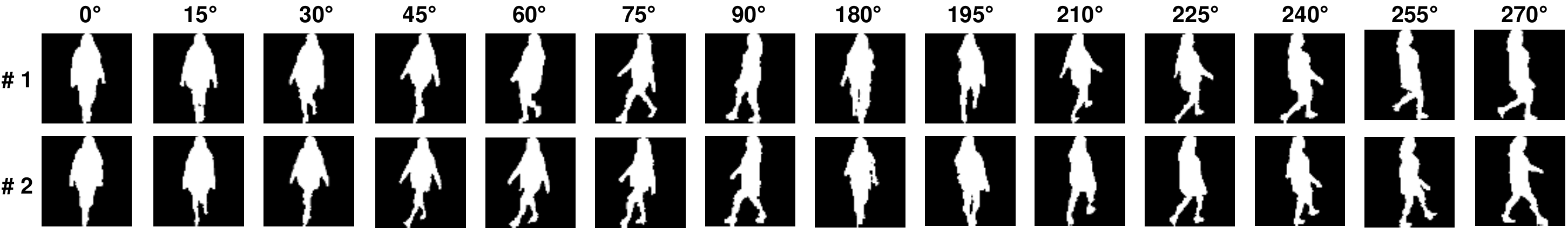}    
    \end{center}
\caption{Example cropped gait silhouettes from one subject in OU-MVLP dataset \cite{MVLP}, captured in 14 different view angles \{$0^\circ$, $15^\circ$, \dots, $90^\circ$\} and \{$180^\circ$, $195^\circ$, \dots, $270^\circ$\} in two sessions.}
\label{fig:mvlp}
\end{figure*}

\section{Experiments}
In this section we present the used datasets and test protocols along with the implementation details and state-of-the-art methods used for benchmarking. We will also discuss the experimental setup used to evaluate the robustness of the benchmarking methods with respect to a number of synthesized occlusions.

\subsection{Datasets}
We have conducted our experiments on two large-scale gait datasets using four different test protocols.

\textbf{CASIA-B dataset} \cite{CASIA}, as one of the most well-known multi-view gait datasets, contains data from 124 subjects captured in three different walking conditions, including normal walking (6 gait sequences per person per view), walking with a coat (2 gait sequences per person per view), and walking with a bag (2 gait sequences per person per view). The gait data has been captured from 11 different angles, ranging from 0$^{\circ}$ to 180$^{\circ}$ (18$^{\circ}$ angle change in each step), as shown in Figure \ref{fig:casia}.

\textbf{OU-MVLP dataset} \cite{MVLP}, is currently the largest gait dataset available, containing data from 10,307 subjects, which have been captured in two different acquisition sessions. This dataset has an almost equal gender distribution with an age range of 2 to 87 years. The gait data has been captured from 14 different angles, ranging from 0$^{\circ}$ to 90$^{\circ}$, and 180$^{\circ}$ to 270$^{\circ}$ with 15$^{\circ}$ angle change in each step, per each subject, as shown in Figure \ref{fig:mvlp}. 

The CASIA-B and OU-MVLP datasets directly provide gait silhouette sequences. In the experiments, we first calculated the gait bounding boxes and aligned the silhouette images. We then cropped the gait silhouettes from the full frames and resized them into a 64$\times$64 pixel image, to be used as inputs to our deep model. This resolution has been proven to be sufficient for accurate gait recognition while maintaining low computational cost \cite{B6}.

\subsection{Benchmarking and Test Protocols}
As described in Section 2, there are a number of deep gait recognition methods whose results have been reported on the same datasets using the same test protocols. These methods are used for our benchmarking study. Apart from these methods, the experiments also include two baseline methods. In the first baseline method denoted as \textit{Global Representation}, GCEM global representations (without splitting) are used as input to a fully connected layer for dimensionality reduction followed by the classifier to perform gait recognition. The second baseline method, denoted as \textit{Subnetwork A}, directly uses the partial representations obtained from the first subnetwork, as illustrated in Figure \ref{fig:2}(A), along with a softmax layer for performing classification. This baseline analysis can reveal the added value of the attentive recurrent learning of partial representation.

For CASIA-B dataset\cite{CASIA}, we have used a test protocol that has frequently been used in the literature \cite{B1, B2, B3, B4, B6, B7, B10}, using the data from the first 74 subjects for training and the the remaining 50 subjects for testing. During testing, the first four gait sequences from the normal walking condition have been used to form the gallery set, where the rest of the sequences form three probe sets. The probe sets include the remaining 2 normal walking sequences (NM), 2 walking sequences with a coat (CL), and 2 walking sequences with a bag (BG), per each subject per each view. The results have been reported for all 11 angles, where the probe samples with identical angles have been excluded from the reference set during testing. Concerning OU-MVLP dataset \cite{MVLP}, we used the protocol available along with the dataset, thus using a pre-determined set of 5153 and 5154 subjects respectively for training and testing sets. Similar to the experiments performed in \cite{B8, B6, B9, B1}, four viewing angles, notably 0$^{\circ}$, 30$^{\circ}$, 60$^{\circ}$, and 90$^{\circ}$, have been considered for cross-view gait recognition.

\subsection{Robustness to Occlusions}
A number of different factors may cause occlusions in gait images in real-word scenarios. For example, occlusions occur frequently when the person of interest is covered by obstacles such as other people in the scene or objects like desks, making it impossible for the camera to capture all the necessary parts of the subject's body.
Another scenario is the occurrence of self-occlusions, where parts of the body that are closer to the camera obstruct the view of other parts. Finally, as gait silhouettes are usually extracted by finding the difference between the body parts and the background, there might be different anomalies within the extracted silhouettes, such as spurious pixels or holes, due to complex backgrounds or motion-related distortions such as pixel displacement. As a result of these potential scenarios, occlusions might be encountered frequently, causing considerable degradation in the performance of gait recognition systems \cite{incomp}.

\begin{figure}[!t]
    \begin{center}
    \includegraphics[width=0.80\linewidth]{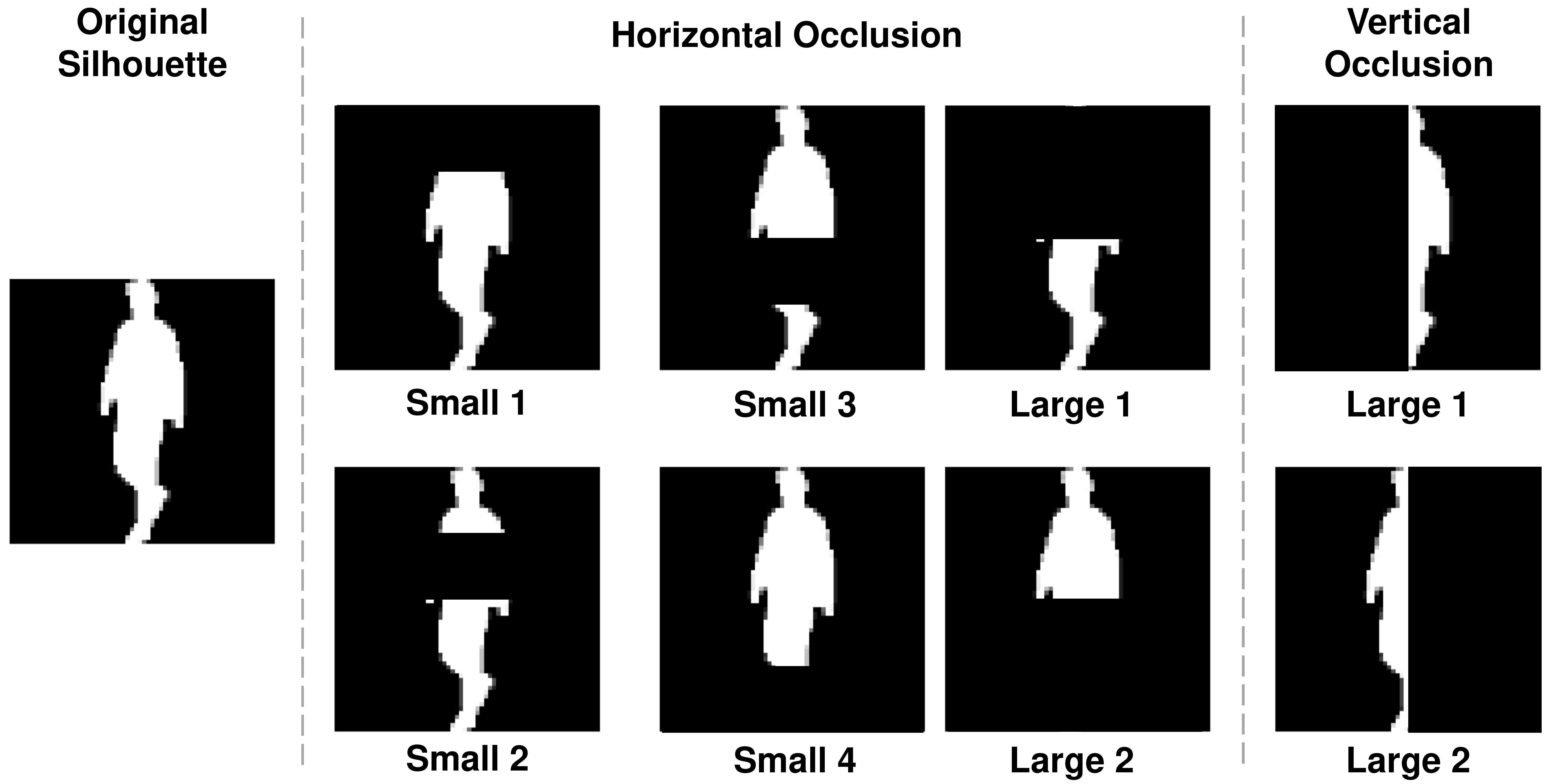}
    \end{center}
\caption{First column: original gait silhouettes; second and third columns: small synthesized horizontal occlusions added to the original gait silhouettes; fourth column: large synthesized horizontal occlusions added to the original silhouettes; and fifth column: large synthesized vertical occlusions added to the the original gait silhouettes.}
\label{fig:Occlusion}
\end{figure}

To test the robustness of our proposed model toward occlusions, we used synthesized occlusions by adding horizontal and vertical bars with the same color as the background to the gait silhouettes. The assumption here is that training has been done with complete (non-occluded) data, while testing includes gallery and probe gait data whose silhouettes contain missing body parts. This type of study has been widely used in the gait recognition literature \cite{incomp, bar1, bar2, bar3}. We have designed two classes of synthesized horizontal occlusions, namely \textit{ii}) small occlusions, where quarters of the gait silhouettes are occluded by the horizontal bars with the width of 16 pixels and step size of 16 pixels, thus creating four occluded synthesized gait silhouettes as shown in the second and third columns of Figure \ref{fig:Occlusion}; and \textit{i}) large occlusions where the top-half and bottom-half of the gait silhouettes are occluded respectively (Figure \ref{fig:Occlusion}, fourth column). Concerning the vertical occlusions, only large occlusions (Figure \ref{fig:Occlusion}, fifth column) have been considered. Small vertical occlusions were not considered as the first and last vertical quarters of the gait silhouettes mainly contain only the background as shown in Figure \ref{fig:Occlusion}.
The ability of the proposed method, as well as the benchmarking methods, are then evaluated in the presence of these occlusions.

\begin{table}[!t]
\centering
\caption{Selected parameter values for our network.}
\setlength
\tabcolsep{4pt}
\begin{tabular}{ l| l| l}
\hline
\textbf{Subnetwork} & \textbf{Parameter} & \textbf{Setting} \\
\hline
\hline
    Subnetwork A & Loss Function & Triplet Loss  \\
      & Margin Value & 0.2 \\
      & Triplet Mining & Batch All \\
      & Learning Rate & 0.0001 \\
      & Batch Size (CASIA-B)& 128 \\
      & Batch Size (OU-MVLP)& 512 \\
      & Number of Epochs (CASIA-B) & 80000\\
      & Number of Epochs (OU-MVLP) & 100000\\
    \hline

    Subnetwork B & GRU Hidden Layer Size  & 128  \\
      & Dropout Rate  & 0.1  \\
      & Loss Function & Cosine Proximity \\
      & Optimizer & Adam \\
      & Metric & Accuracy \\
      & Learning Rate & 0.0001 \\
      & Batch Size (CASIA-B) & 220 \\
      & Batch Size (OU-MVLP) & 500 \\
      & Number of Epochs (CASIA-B) & 1500\\
      & Number of Epochs (OU-MVLP) & 5000\\

    \hline
\end{tabular}
\label{tab: parameter}
\end{table}

\subsection{Implementation Details}
The parameters' values used to obtain the best results for both subnetworks are summarized in Table \ref{tab: parameter}. It should be mentioned that due to the different sizes of the two datasets, different batch sizes were used during training of the two subnetworks. The number of training epochs was selected based on maximum validation performance. 
Our model has been implemented using TensorFlow with Keras backend and has been trained with four NVIDIA RTX 2080 Ti GPUs.

\begin{figure}[t!]
\centering
\includegraphics[width=1\columnwidth]{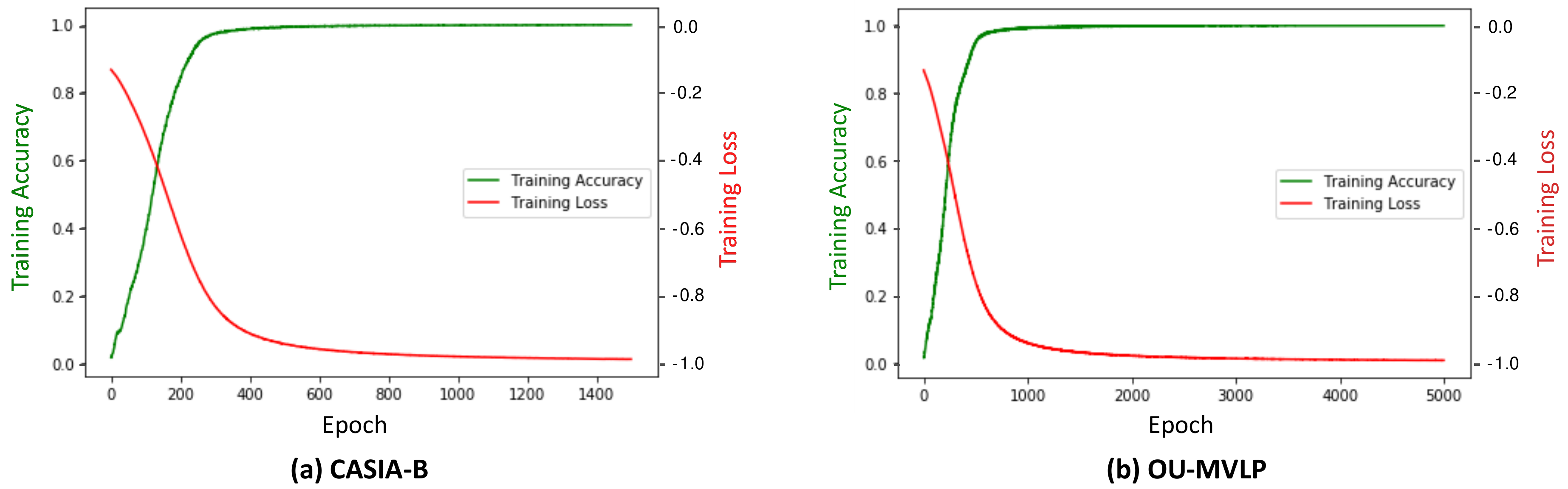}
\caption{Accuracy and loss curves for our proposed solution.}
\label{fig:loss}
\end{figure}

\begin{table*}[!t]
\centering
\caption{Gait recognition results on the CASIA-B dataset under normal (NM) walking for the proposed and other benchmarking methods.}
 \setlength\tabcolsep{5pt}
\begin{tabular}{ l l l l | l l l l l l l l l l l | l}
\hline
\multicolumn{4}{ c |}{\textbf{Method}} & \multicolumn{12}{ c }{\textbf{View}}  \\ 
\hline
	\textbf{Name} & \textbf{Reference} & \textbf{Year} & \textbf{Venue} & \textbf{0$^{\circ}$} & \textbf{18$^{\circ}$} & \textbf{36$^{\circ}$} & \textbf{54$^{\circ}$} & \textbf{72$^{\circ}$} & \textbf{90$^{\circ}$} & \textbf{108$^{\circ}$} & \textbf{126$^{\circ}$} & \textbf{144$^{\circ}$} & \textbf{162$^{\circ}$} & \textbf{180$^{\circ}$} & \textbf{Mean $\pm$ SD} \\ \hline\hline
	CNN-3D & ~\cite{B6} & 2017 & \textit{T-PAMI} & 87.1 & 93.2 & 97.0 & 94.6 & 90.2 & 88.3 & 91.1 & 93.8 & 96.5 & 96.0 & 85.7 & 92.1 $\pm$ 4.0 \\ 
	CNN-Ensem. & ~\cite{B6} & 2017 & \textit{T-PAMI} & 88.7 & 95.1 & 98.2 & 96.4 & 94.1 & 91.5 & 93.9 & 97.5 & 98.4 & 95.8 & 85.6 & 94.1 $\pm$ 4.0 \\ 
	MGAN & ~\cite{B7} & 2019 & \textit{T-IFS} & - & - & - & 84.2 & - & 72.3 & - & 83.0 & - & - & - & 79.8  $\pm$ 6.5\\ 
	EV-Gait & ~\cite{B2} & 2019 & \textit{CVPR} & 77.3 & 89.3 & 94.0 & 91.8 & 92.3 & 96.2 & 91.8 & 91.8 & 91.4 & 87.8 & 85.7 & 89.9  $\pm$ 5.1\\ 
	Gait-Joint & ~\cite{B10} & 2019 & \textit{PR} & 75.6 & 91.3 & 91.2 & 92.9 & 92.5 & 91.0 & 91.8 & 93.8 & 92.9 & 94.1 & 81.9 & 89.9 $\pm$ 5.8\\
	GaitSet & ~\cite{B1} & 2019 & \textit{AAAI} & 90.8 & 97.9 & 99.4 & 96.9 & \textbf{93.6} & 91.7 & 95.0 & 97.8 & \textbf{98.9} & \textbf{96.8} & 85.8 & 95.0  $\pm$ 4.2\\ 
	GaitNet &~\cite{B3} & 2019 & \textit{CVPR} & 91.2 & - & - & 95.6 & - & 92.6 & - & 96.0 & - & - & - & 93.9  $\pm$ 2.3\\
	GaitNet-2 & ~\cite{B4} & 2020 & \textit{arXiv} & 93.1 & 92.6 & 90.8 & 92.4 & 87.6 & \textbf{95.1} & 94.2 & 95.8 & 92.6 & 90.4 & \textbf{90.2} & 92.3  $\pm$ 2.4 \\ 
	\hline
	Global Rep. & Baseline 1 & - & - & 74.0 & 84.5 & 89.9 & 86.3 & 79.7 & 77.4 & 83.0 & 86.6 & 85.6 & 80.3 & 69.2 & 81.5 $\pm$ 6.1 \\
	{Subnetwork A} & Baseline 2 & - & - & 89.00 & 96.9 & \textbf{99.5} & 96.7 & 91.4 & 88.6 & 94.5 & 97.4 & 97.7 & 94.00 & 83.3 & 93.5 $\pm$ 4.9 \\
	\textbf{Ours} & - & - & - & \textbf{91.1} & \textbf{98.0} & 99.4 & \textbf{98.2} & {93.2} & 91.9 & \textbf{95.2} & \textbf{98.3} & 98.4 & 95.7 & 87.5 & \textbf{95.2 $\pm$ 3.8}  \\ 
\hline
\end{tabular}
\label{tab2}
\end{table*}

\begin{table*}[!t]
\centering
\caption{Gait recognition results on the CASIA-B dataset with different carried bags (BG), for the proposed and other  benchmarking methods.}
\setlength\tabcolsep{5pt}
\begin{tabular}{ l l l l | l l l l l l l l l l l | l}
\hline
\multicolumn{4}{ c |}{\textbf{Method}} & \multicolumn{12}{ c }{\textbf{View}}  \\
\hline
	\textbf{Name} & \textbf{Reference} & \textbf{Year} & \textbf{Venue} & \textbf{0$^{\circ}$} & \textbf{18$^{\circ}$} & \textbf{36$^{\circ}$} & \textbf{54$^{\circ}$} & \textbf{72$^{\circ}$} & \textbf{90$^{\circ}$} & \textbf{108$^{\circ}$} & \textbf{126$^{\circ}$} & \textbf{144$^{\circ}$} & \textbf{162$^{\circ}$} & \textbf{180$^{\circ}$} & \textbf{Mean $\pm$ SD} \\ \hline\hline
	CNN-LB & ~\cite{B6} & 2017 & \textit{T-PAMI} & 64.2 & 80.6 & 82.7 & 76.9 & 64.8 & 63.1 & 68 & 76.9 & 82.2 & 75.4 & 61.3 & 72.4 $\pm$ 8.4 \\ 
	MGAN & ~\cite{B7} & 2019 & \textit{T-IFS} & 48.5 & 58.5 & 59.7 & 58 & 53.7 & 49.8 & 54 & 61.3 & 59.5 & 55.9 & 43.1 & 54.7 $\pm$ 5.6 \\ 
	GaitSet & ~\cite{B1} & 2019 & \textit{AAAI} & 83.8 & 91.2 & 91.8 & 88.8 & 83.3 & 81.0 & 84.1 & 90.0 & 92.2 & \textbf{94.4} & 79.0 & 87.2  $\pm$ 5.2\\ 
	GaitNet & ~\cite{B3} & 2019 & \textit{CVPR} & 83 & - & - & 86 & - & 74.8 & - & 85.8 & - & - & - & 82.6  $\pm$ 5.2\\ 
	\hline
	Global Rep. & Baseline 1 & - & - & 64.7 & 73.8 & 80.1 & 75.6 & 67.9 & 64.9 & 71.0 & 77.8 & 79.4 & 77.0 & 66.6 & 72.6 $\pm$ 5.9 \\
	{Subnetwork A} & Baseline 2 & - & - & 81.6 & 91.7 & 91.6 & 89.1 & 82.1 & 80.00 & 82.9 & 90.8 & 92.7 & 91.6 & 77.9 & 86.5  $\pm$ 5.6 \\
	\textbf{Ours} & - & - & - & \textbf{86.0} & \textbf{93.3} & \textbf{95.1} & \textbf{92.1} & \textbf{88.0} & \textbf{82.3} &  \textbf{87.0}& \textbf{94.2} & \textbf{95.9} & 90.7 & \textbf{82.4} & \textbf{89.7 $\pm$ 4.9} \\
\hline
\end{tabular}
\label{tab3}
\end{table*}

\begin{table*}[!t]
\centering
\caption{Gait recognition results on CASIA-B dataset with different clothing (CL) for the proposed and other benchmarking methods.}
 \setlength\tabcolsep{5pt}
\begin{tabular}{ l l l l | l l l l l l l l l l l | l}
\hline
\multicolumn{4}{ c |}{\textbf{Method}} & \multicolumn{12}{ c }{\textbf{View}}  \\ 
\hline
	\textbf{Name} & \textbf{Reference} & \textbf{Year} & \textbf{Venue} & \textbf{0$^{\circ}$} & \textbf{18$^{\circ}$} & \textbf{36$^{\circ}$} & \textbf{54$^{\circ}$} & \textbf{72$^{\circ}$} & \textbf{90$^{\circ}$} & \textbf{108$^{\circ}$} & \textbf{126$^{\circ}$} & \textbf{144$^{\circ}$} & \textbf{162$^{\circ}$} & \textbf{180$^{\circ}$} & \textbf{Mean $\pm$ SD} \\ \hline\hline
	CNN-LB & ~\cite{B6} & 2017 & \textit{T-PAMI} & 37.7 & 57.2 & 66.6 & 61.1 & 55.2 & 54.6 & 55.2 & 59.1 & 58.9 & 48.8 & 39.4 & 54.0  $\pm$ 8.9 \\ 
	MGAN & ~\cite{B7} & 2019 & \textit{T-IFS} & 23.1 & 34.5 & 36.3 & 33.3 & 32.9 & 32.7 & 34.2 & 37.6 & 33.7 & 26.7 & 21.0 & 31.5  $\pm$ 5.4\\ 
	GaitSet & ~\cite{B1} & 2019 & \textit{AAAI} & 61.4 & 75.4 & 80.7 & 77.3 & 72.1 & 70.1 & 71.5 & 73.5 & 73.5 & 68.4 & 50.0 & 70.4 $\pm$ 8.4\\ 
	GaitNet & ~\cite{B3} & 2019 & \textit{CVPR} & 42.1 & - & - & 70.7 & - & 70.6 & - & 69.4 & - & - & - & 63.2  $\pm$ 14.1\\ 
	\hline
	Global Rep. & Baseline 1 & - & - & 55.0 & 62.4 & 66.0 & 61.1 & 55.9 & 54.4 & 59.4 & 60.8 & 62.8 & 52.7 & 44.6 & 57.7 $\pm$ 6.0 \\
	{Subnetwork A} & Baseline 2 & - & - & 57.2 & 76.1 & 80.9 & 77.2 & 72.3 & 70.2 & 73.0 & 72.6 & 75.0 & 69.6 & 50.9 & 70.4 $\pm$ 8.8 \\
	\textbf{Ours} & - & -  & - & \textbf{65.8} & \textbf{80.7} & \textbf{82.5} & \textbf{81.1} & \textbf{72.7} & \textbf{71.5} & \textbf{74.3} & \textbf{74.6} & \textbf{78.7} & \textbf{75.8} & \textbf{64.4} & \textbf{74.7 $\pm$ 5.9}\\  
\hline
\end{tabular}
\label{tab4}
\end{table*}

\begin{table*}[!t]
\centering
\caption{Gait recognition results on the OU-MVLP dataset for the proposed and other benchmarking methods.}
\begin{tabular}{ l l l l | l l l l | l }
\hline
\multicolumn{4}{ c |}{\textbf{Method}} & \multicolumn{5}{ c }{\textbf{View}}  \\ 
\hline
	\textbf{Name} & \textbf{Reference} & \textbf{Year} & \textbf{Venue} &  \textbf{0$^{\circ}$} & \textbf{30$^{\circ}$} & \textbf{60$^{\circ}$} & \textbf{90$^{\circ}$} & \textbf{Mean $\pm$ SD} \\ \hline\hline
	GEI NET & ~\cite{B8} & 2016 &\textit{ICB} & 15.7 & 41.0 & 39.7 & 39.5 & 34.0 $\pm$ 12.2 \\ 
	CNN-LB & ~\cite{B6} & 2017 &\textit{T-PAMI} & 14.2 & 32.7 & 32.3 & 34.6 & 28.5  $\pm$ 9.5\\ 
	DigGAN & ~\cite{B9} & 2018 & arXiv & 30.8 & 43.6 & 41.3 & 42.5 & 39.6  $\pm$ 5.9 \\ 
	GaitSet & ~\cite{B1} & 2019 & \textit{AAAI} & 77.7 & 86.9 & 85.3 & 83.5 & 83.4  $\pm$ 4.1\\ 
	\hline
	Global Rep. & Baseline 1 & - & - & 68.9 & 82.3 & 82.1  &  81.7 &  78.8 $\pm$ 6.6 \\
	{Subnetwork A} & Baseline 2 & - & - & 74.7  & 84.4  &  83.7  &   82.2 &  81.3  $\pm$ 4.5  \\
	\textbf{Ours} & - & - & - & \textbf{78.5} & \textbf{87.5} & \textbf{85.8} & \textbf{85.4} & \textbf{84.3 $\pm$ 4.0} \\
\hline
\end{tabular}
\label{tab5}
\end{table*}

\section{Results and Analysis}
This section presents the cross-view recognition performance and analysis for the proposed and benchmarking gait recognition solutions. We also study the robustness against different occlusions. Finally, ablation experiments are performed to investigate the effect of the individual components of the proposed network on the overall recognition performance.

\subsection{Performance}
Tables \ref{tab2}, \ref{tab3}, and \ref{tab4} show the comparison between our model and state-of-the-art methods for CASIA-B test protocols, including normal walking (NM), walking with a bag (BG), and walking with different clothing (CL) such as a coat, respectively. Additionally, Table \ref{tab5} presents the comparison between our model and state-of-the-art methods for the OU-MVLP dataset. In order to perform a fair comparison, we strictly followed the respective protocols for cross-view gait recognition in the benchmarking methods as presented in Section 4.2. The reported results, in term of rank 1 identification, are averaged across all gallery views except identical angles. It is worth noting that the OU-MVLP dataset has been collected fairly recently, thus the number of benchmarking methods for this dataset is naturally lower than that for CASIA-B.

The results (Tables \ref{tab2}, \ref{tab3}, \ref{tab4} and \ref{tab5}) show that our model achieves the best recognition performance for most of the view angles/test protocols considered, setting new state-of-the-art average values for both datasets using all four test protocols. The results also show the significant added value of recurrently learned partial gait representation, in the context of our model, when compared to the baseline methods using partial representations (without attentive recurrent learning) and global representations. The superiority of our model is more evident under the more challenging appearance variations, i.e., walking when wearing different clothing (Table \ref{tab3}) and carrying bags (Table \ref{tab4}). This is due to the ability of the attentive recurrent component, which adapts the model to variations by focusing on the most informative spatiotemporal parts of the embedding, thus learning more robust representations under appearance changes.

\begin{table*}[!t]
\centering
\caption{Gait recognition results for the proposed and two other benchmarking methods in the the presence of six different occlusions.}
\begin{tabular}{ l | l | l l l l l l l l l l l | l}
\hline

\hline
	\textbf{Occlusion} & \textbf{Method} & \textbf{0$^{\circ}$} & \textbf{18$^{\circ}$} & \textbf{36$^{\circ}$} & \textbf{54$^{\circ}$} & \textbf{72$^{\circ}$} & \textbf{90$^{\circ}$} & \textbf{108$^{\circ}$} & \textbf{126$^{\circ}$} & \textbf{144$^{\circ}$} & \textbf{162$^{\circ}$} & \textbf{180$^{\circ}$} & \textbf{Mean $\pm$ SD} \\ 
	\hline\hline
    Large, 1 & Global Rep.& 42.3 & 56.1 & 65.4 & 64.0 & 56.1 & 53.1 & 59.9 & 65.1 & 59.6 & 48.3 & 37.2 & 55.2 $\pm$ 9.3 \\ 
    Horizontal & GaitSet ~\cite{B1} & 67.6 & 87.5 & 91.6 & 88.7 & 82.5 & 78.3 & 82.7 & 87.2 & 90.9 & 82.9 & 62.1 & 82.0 $\pm$ 9.4  \\ 	
    & Ours & 73.6 & 83.9 & 89.3 & 87.5 & 80.2 & 78.1 & 83.8 & 88.8 & 89.8 & 86.9 & 70.9 & \textbf{83.0} $\pm$ 6.5  \\ 	
    \hline
    Large, 2 & Global Rep.& 44.9 & 52.6 & 60.7 & 59.4 & 55.9 & 54.2 & 60.0 & 62.6 & 60.1 & 50.2 & 42.1 & 54.8 $\pm$ 6.8  \\ 
    Horizontal & GaitSet ~\cite{B1} & 61.9 & 75.8 & 83.6 & 81.2 & 74.7 & 70.8 & 75.4 & 81.9 & 78.3 & 73.6 & 60.8 & 74.4 $\pm$ 7.5  \\ 	
    & Ours & 64.5 & 79.4 & 84.0 & 82.6 & 76.7 & 73.4 & 77.7 & 83.2 & 82.3 & 75.7 & 65.5 & \textbf{76.8} $\pm$ 6.8 \\ 	
    \hline
    Small, 1 & Global Rep.& 45.5 & 64.1& 69.8 & 69.8 & 61.1 & 61.1 & 65.9 & 72.4 & 66.6 & 53.6 & 45.4 & 61.4 $\pm$ 9.4 \\ 
    Horizontal& GaitSet ~\cite{B1} & 84.7 & 94.9 & 97.3 & 95.8 & 91.0 & 89.7 & 91.6 & 94.7 & 95.7 & 93.0 & 76.5 & 91.3 $\pm$ 6.1 \\ 	
    & Ours & 87.1 & 94.0 & 94.9 & 92.5 & 88.2 & 86.6 & 89.0 & 94.4 & 95.8 & 95.8 & 89.1 & \textbf{91.6} $\pm$ 3.7\\ 	
    \hline
    Small, 2 & Global Rep.& 49.1 & 65.2 & 70.3 & 67.1 & 57.5 & 59.8 & 63.9 & 71.9 & 70.8 & 64.0 & 50.8 & 62.5 $\pm$ 7.9  \\ 
    Horizontal& GaitSet ~\cite{B1} & 76.8 & 91.6 & 96.4 & 93.0 & 86.2 & 82.0 & 85.8 & 92.6 & 95.0 & 87.9 & 72.9 & 87.3 $\pm$ 7.6  \\ 	
    & Ours & 82.7 & 90.3 & 94.6 & 92.4 & 83.5 & 81.5 & 84.6 & 91.0 & 94.5 & 90.4 & 82.1 & \textbf{88.0} $\pm$ 5.1 \\ 	
    \hline
    Small, 3 & Global Rep.& 48.9 & 61.1 & 68.7 & 69.5 & 66.0 & 63.2 & 67.6 & 71.6 & 66.0 & 56.2 & 51.2 & 62.7 $\pm$ 7.6  \\ 
    Horizontal& GaitSet ~\cite{B1} & 74.8 & 85.6 & 91.5 & 87.9 & 81.9 & 78.2 & 83.0 & 89.3 & 88.3 & 82.5 & 71.3 & 83.1 $\pm$ 6.3 \\ 	
    & Ours & 73.9 & 86.5 & 92.9 & 88.0 & 82.8 & 81.0 & 83.5 & 91.3 & 92.4 & 83.5 & 69.4 & \textbf{84.1} $\pm$ 7.4  \\ 	
    \hline
    Small, 4 & Global Rep.& 45.8 & 56.7 & 65.7 & 66.1 & 61.6 & 53.5 & 62.2 & 69.5 & 65.4 & 54.7 & 47.6 & 59.0 $\pm$ 7.9   \\ 
    Horizontal& GaitSet ~\cite{B1} & 79.5 & 90.4 & 96.5 & 94.2 & 86.7 & 85.7 & 86.7 & 92.8 & 93.0 & 88.5 & 79.4 & \textbf{88.5} $\pm$ 5.6 \\ 	
    & Ours & 77.9 & 92.9 & 94.1 & 92.4 & 88.7 & 83.2 & 87.6 & 91.5 & 93.2 & 90.1 & 81.5 & \textbf{88.5} $\pm$ 5.4  \\ 	
    \hline
    {Large, 1} & Global Rep.& 11.3  & 14.6  & 14.4  & 18.1  & 25.7  & 20.1  & 24.5  & 26.5  & 21.9  & 14.1  & 11.9  & 18.5  $\pm$ 5.6 \\ 
    {Vertical} & GaitSet ~\cite{B1} &  27.6 & 38.9  &  51.2 & 53.0  & 54.2  & 45.5  & 49.5  & 55.9  &  52.2  & 31.2  & 18.4  &  43.4 $\pm$ 12.6 \\ 	
    & Ours & 35.8  &  46.0 & 57.4  & 57.5  &  55.8 & 46.2  & 47.9  & 55.0  &  53.8 & 38.3  & 28.2  & \textbf{47.4}  $\pm$ 9.8 \\ 	
    \hline
    {Large, 2} & Global Rep.&  07.7  & 20.9  & 21.2  & 19.4  &  19.5 & 19.5  & 22.9  & 23.5  & 20.0  & 16.9  & 10.4  & 18.4 $\pm$ 5.0 \\ 
    {Vertical} & GaitSet ~\cite{B1} & 21.6  & 42.4  & 55.2  & 57.1  & 51.5  & 49.0  & 55.7  & 59.1  & 56.8  & 45.5  & 24.4  &  47.0 $\pm$ 13.0 \\ 	
    & Ours & 30.0  & 46.7  &  54.3 & 51.2 & 50.3  &  47.4 & 58.3  & 60.6  & 61.7  &  48.0 &  25.4  &   \textbf{48.5}  $\pm$ 11.6 \\ 	
\hline
\end{tabular}
\label{occ:NM}
\end{table*}

Similar to the mentioned benchmarking methods, our proposed model achieves less impressive results when facing gait data captured at 0$^{\circ}$, 90$^{\circ}$, and 180$^{\circ}$. This is due to the fact that the appearance of the gait silhouettes substantially changes in extreme angles, i.e., 0$^{\circ}$ and 180$^{\circ}$, where the camera lens is parallel to the walking direction and thus some gait information like stride are occluded. Additionally, some other useful gait information, such as body swing, may be invisible in the 90$^{\circ}$ angle when the camera lens is orthogonal to the walking direction. Nevertheless, the results show that our method performs better than the benchmarking methods in most of these difficult cases, revealing that our method learns more discriminative features for extreme viewing angle changes. Our method also provides impressive results when facing gait data captured at intermediate angles, e.g., 36$^{\circ}$, 54$^{\circ}$, 126$^{\circ}$, and 144$^{\circ}$, where useful gait information such as body shape and walking postures are visible.

Another interesting observation is that our model generalizes well to different number of subjects and training samples, as well as imaging setups. CASIA-B and OU-MVLP datasets have been collected under different acquisition conditions and include 259,013 and 13,680 gait sequences from 125 and 10307 individuals respectively. Nonetheless, our model performs better than the benchmarking methods on both datasets fairly consistently.

Finally, it is worth noting that our model offers very compact gait representations with a dimension of 4,096 when compared to the best performing benchmarking method, GaitSet \cite{B1}, whose feature size is 15,872. This can simplify the classification, retrieval and transmission of the extracted gait representations.

\subsection{Robustness Results}
The performance of our model and two other benchmarking methods, including global representations and Gaitset \cite{B1} (as the best performing benchmarking method), has been evaluated using silhouettes with synthesized occlusions as discussed in Section 4.3. The analysis has been performed using CASIA normal (NM) gait samples to study the effects of occlusions rather in isolation from other appearance variations namely carrying a bag and wearing different clothing. In this context, Table \ref{occ:NM} presents the results for all the available views. As the results show, our method consistently achieves the best average performance, compared to the competing methods, for the six occlusion types considered. For the horizontal occlusions, the results show that upper parts of the gait silhouettes contain less informative information for the gait recognition process, compared to the lower parts. Moreover, we observe that the vertical occlusions cause considerable performance degradation compared to the horizontal ones. This maybe be due to the fact that considerable \textit{viewpoint-specific} spatiotemporal information is lost in the presence of vertical occlusions, resulting in higher degradation compared to horizontal occlusions that affect information that is less viewpoint-specific.

\begin{figure}[t]
    \begin{center}
    \includegraphics[width=1\linewidth]{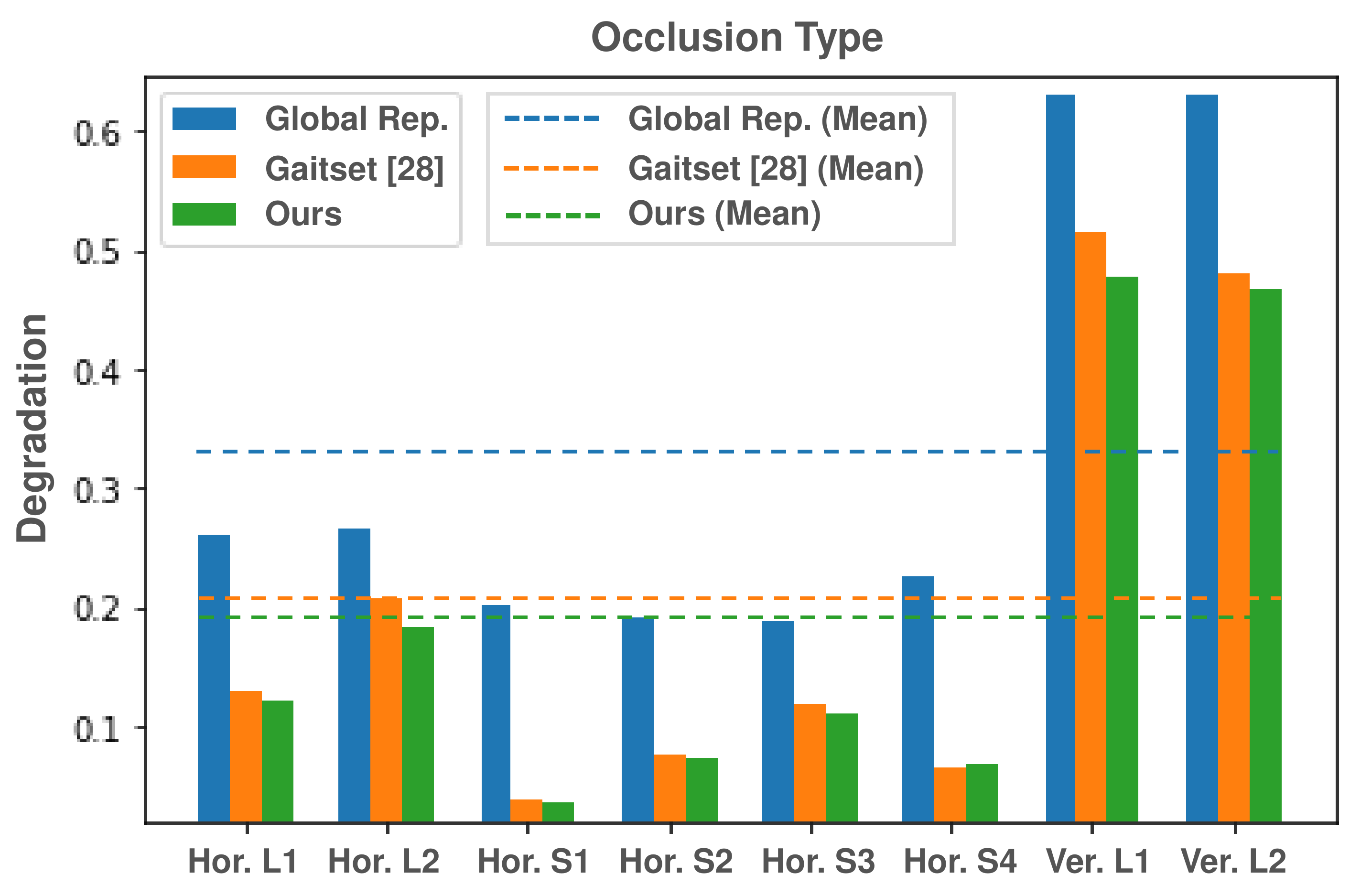} 
    \end{center}
\caption{Performance degradation versus occlusion type.}
\label{fig:rob}
\end{figure}

To further analyze the results, Figure \ref{fig:rob} illustrates the performance \textit{degradation} of each method in the presence of the different types of horizontal (Hor.) and vertical (Ver.) occlusions. This figure reveals that our model is considerably more robust than the baseline global representations approach due to it's ability to adapt by focusing on informative segments of the input frames. Our model also delivers more robust results compared to the Gaitset \cite{B1} technique by attentively exploiting the relations between partial feature embeddings. This performance enhancement is interestingly more pronounced in the case of the larger occlusions.

\begin{figure*}[!t]
\centering
\includegraphics[width=0.8\linewidth]{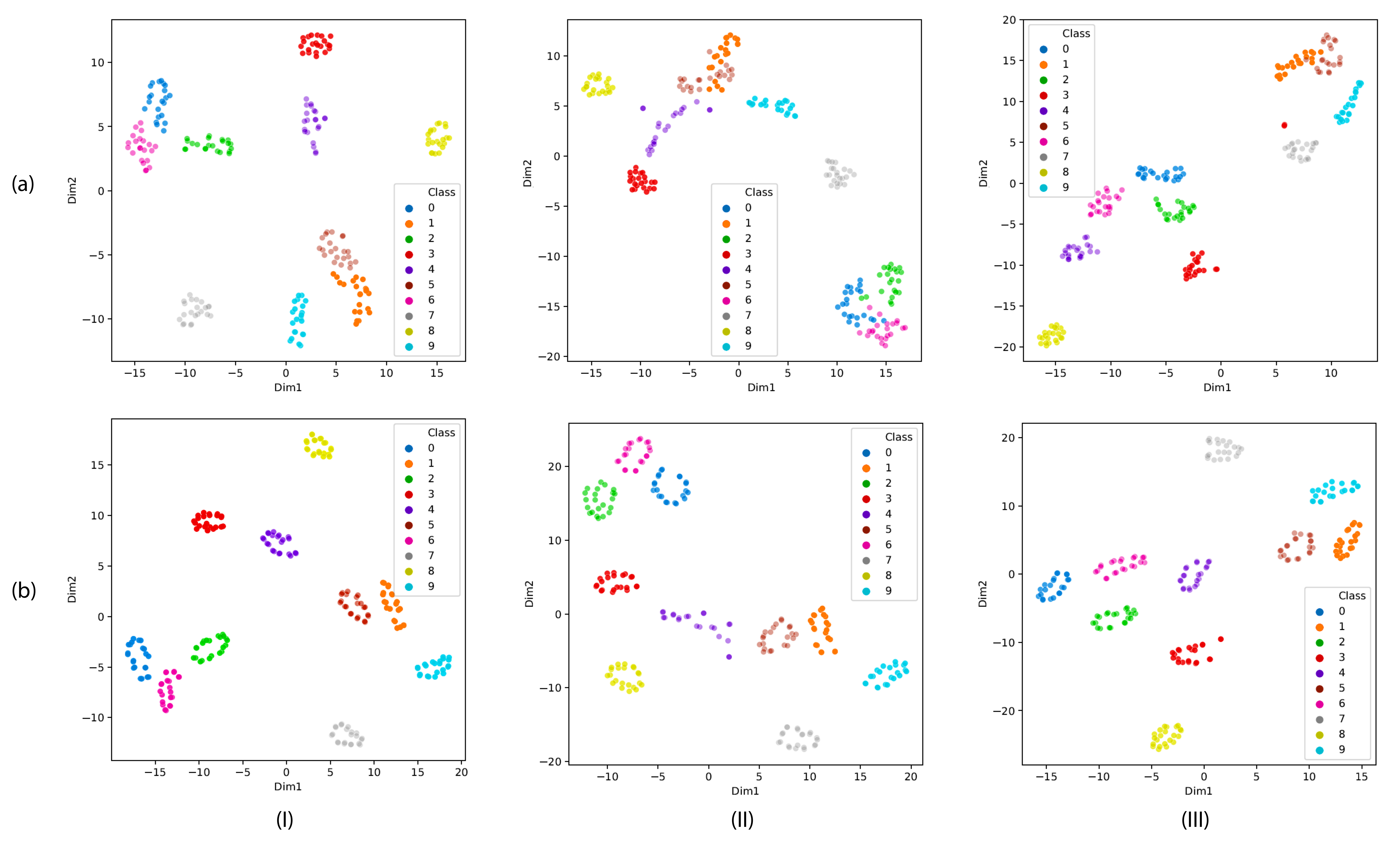}
\caption{t-SNE visualization of the feature spaces using (a) global representations and (b) our method with (I) NM, (II) BG, and (III) CL data.}
\label{fig:tSne}
\end{figure*}

\subsection{Feature Space Exploration}
In order to study the impact of global and partial strategies for representation learning, t-Distributed Stochastic Neighbor Embedding (t-SNE) \cite{tSNE} is used, summarizing the high dimensional feature space in a two-dimensional space. Figure \ref{fig:tSne} illustrates the features that feed the classifier, when \textit{i}) global representations are used, thus using the GCEM features (without splitting) as input to a fully connected layer for dimensionality reduction (Figure \ref{fig:tSne} row a); and \textit{ii}) our model is used to learn the relations between partial representations (Figure \ref{fig:tSne} row b). To make this visualisation more legible, the t-SNE analysis is only performed for the first 10 subjects available in CASIA-B dataset, respectively for NM (Figure \ref{fig:tSne} column I), BG (Figure \ref{fig:tSne} column II), and CL (Figure \ref{fig:tSne} column III) test data.

The first row of Figure \ref{fig:tSne} clearly shows that the global representations could not form separate clusters for some of the test samples. Even in the case of the less challenging NM test protocol, a number of samples can be observed whose data points are mixed with each other.
The results presented in the second row of Figure \ref{fig:tSne}, however, show that our model forms dense and more effective clusters whose data points are distributed more closely around their respective centroids. This observation indicates that in this feature space, the subjects are more separable, validating the use of a recurrent network to learn partial representations in our proposed model.

\subsection{Ablation Experiments}
In order to understand the contribution of each layer of our network towards the final performance, we have performed a number of ablation experiments. Table \ref{tab6} presents the performance of our model when different layers have been systematically removed from the network, one at a time. The last row of the table shows the results of the complete model when all the layers have been included.

\subsubsection{Impact of Splitting} 
We started by removing splitting layer, thus considering GCEM as a single bin which is equivalent to a \textit{global representation}. Naturally, the BGRU would have to be reduced to a single cell with an attention weight of 1 to accommodate this experiment.
The results show that the performance dropped to 70.6\% when considering a single global representation.

\subsubsection{Impact of the Fully Connected Layer} 
Next, we removed the fully connected layer and used the convolutional bins directly as inputs to the BGRU layer. This change not only considerably increased the computational complexity of the BGRU layer due to the higher dimensionality of the input, but also decreased the recognition performance to 63.3\%. In the absence of the fully connected layer, the high-dimension bins would most likely require a deeper network in order to properly learn the larger feature space.

\subsubsection{Impact of the BGRU Layer} 
By removing the BGRU layer, we directly learned the attention weights form the fully connected features. The results show that the performance dropped to 80.6\% as the relations between the partial features, that are likely to contain person-specific information, can not be learned. 

\subsubsection{Impact of the Attention Layer} 
The attention layer was then removed, allowing the recurrently learned GCEM bins to maintain similar importance towards the final goal. The results show than the performance slightly drops when removing this layer. As illustrated in Table \ref{tab6}, the added value of the attention layer is more evident in scenarios where different clothing and carrying conditions are encountered.

\subsubsection{Impact of the Number of Split Bins} 
\begin{figure}[t]
    \begin{center}
    \includegraphics[width=1\linewidth]{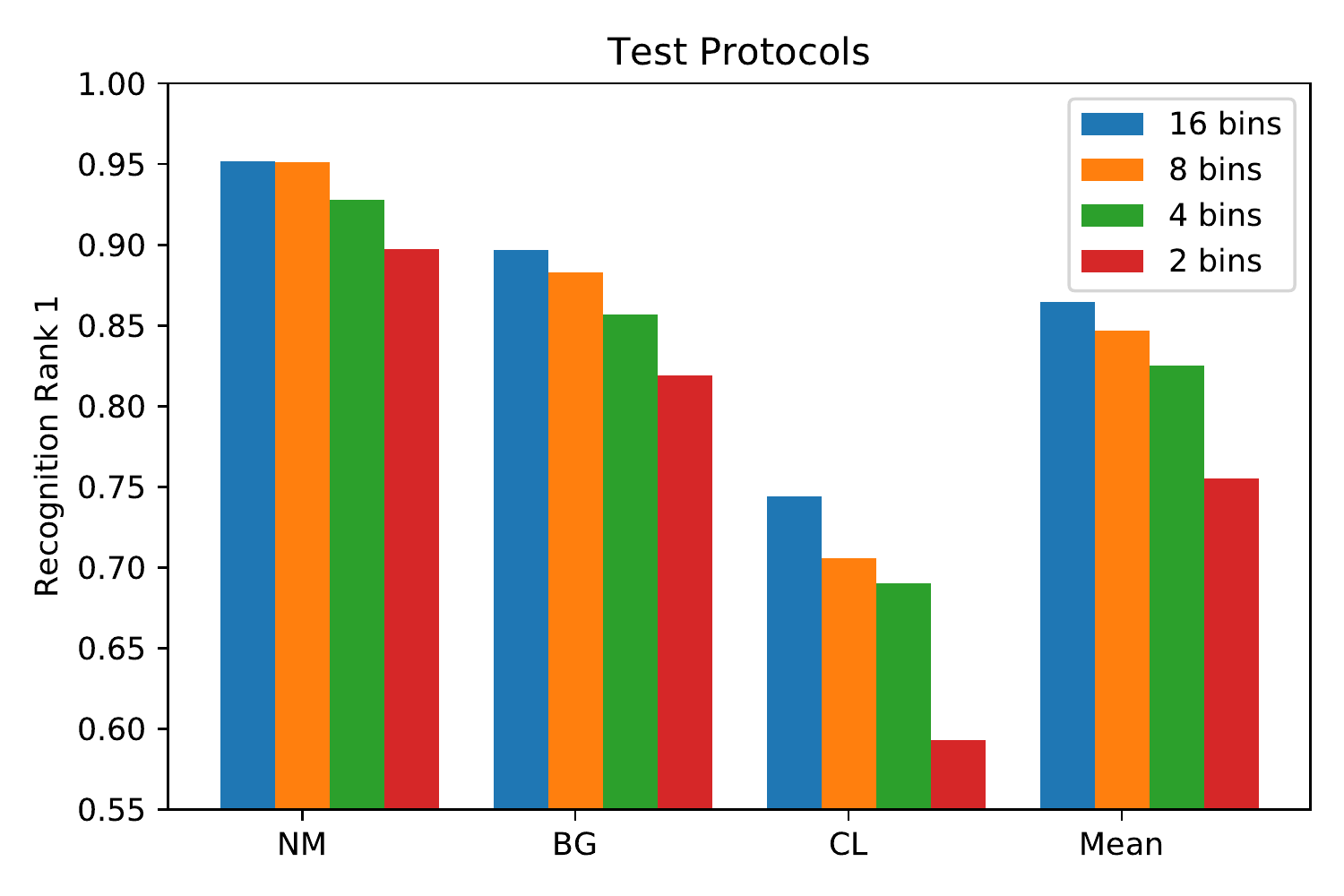} 
    \end{center}
\caption{Rank-1 identification versus number of bins.}
\label{fig:6}
\end{figure}

As discussed in Section 3.4.3, different number of bins can be considered when horizontally splitting the GCEM. Figure \ref{fig:6} illustrates the impact of different number of bins on the gait recognition performance. The results justify the selection of 16 as the number of bins (the maximum possible number of bins) in our model.

\begin{table}[!t]
\centering
\centering
\caption{Ablation study on the CASIA-B dataset.}
\begin{tabular}{ l | l| l| l| l| l| l| l}
\hline
 \multicolumn{1}{ c |}{\textbf{Splitting}}& \multicolumn{1}{ c |}{\textbf{FC}}& \multicolumn{1}{ c| }{\textbf{BGRU}}& \multicolumn{1}{ c| }{\textbf{Att.}}& \multicolumn{4}{ c }{\textbf{Protocol}}  \\ 
\hline
    & & & & NM & BG & CL & Mean\\ \hline\hline
	\xmark & \cmark & \cmark& \cmark & 81.5 & 72.6 & 57.7 & 70.6\\
    \cmark & \xmark & \cmark& \cmark & 91.4 & 71.8 & 27.5 & 63.6\\
    \cmark & \cmark & \xmark& \cmark & 93.9 & 86.7 & 71.1 & 83.9\\
	\cmark & \cmark & \cmark& \xmark & 95.0 & 89.3 & 73.8 & 86.0 \\
	\cmark & \cmark & \cmark& \cmark & \textbf{95.2} & \textbf{89.7} & \textbf{74.4} & \textbf{86.4}\\
    \hline
\end{tabular}
\label{tab6}
\end{table}

\section{Conclusion and Future Work}
We proposed a novel method for learning robust view-invariant person-specific gait features. Our proposed deep network learns partial representations from gait convolutional energy maps to be then learned using an attentive recurrent model. In this context, our model adopted a BGRU network to exploit the relations between the partial representations. We then used an attention mechanism to selectively focus on important recurrently learned partial representations as identity information in different scenarios may lie in different representations. Our method has been extensively tested on two large-scale CASIA-B and OU-MVLP gait datasets using four different test protocols. Additionally, we used synthesized occlusions by adding horizontal and vertical bars with the same color as the background to study the robustness of our method. The obtained recognition results showed the superiority of our model over a number of state-of-the-art methods. 
Finally, an ablation study highlighted the contribution of each component in our network.

In future work, we will investigate the effects of other strategies to extract partial representations. In this context, the temporal templates can be adaptively split into gait patches with different scales, rather than horizontal bins. In addition, external data from other datasets will be used to train a more generalized model that could be then applied to new unseen data.

\bibliographystyle{ieeetran}
\bibliography{References}

\begin{IEEEbiography}[
{
\includegraphics[width=1in,height=1.21in,clip,keepaspectratio]{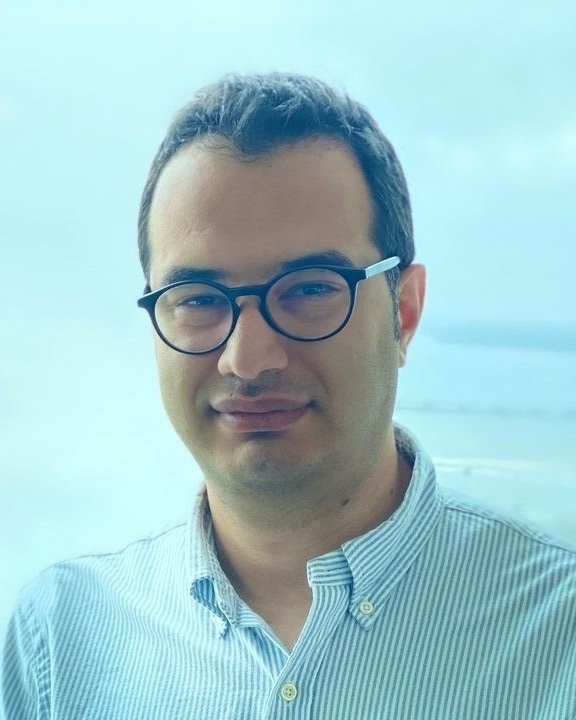}
}
]
{Alireza Sepas-Moghaddam}
received the B.Sc. and M.Sc. (first class honors) degrees in Computer Engineering in 2007 and 2010, respectively. From 2011 to 2015, he was with the Shamsipour Technical University, Tehran, Iran, as a lecturer. In 2015, he joined Instituto Superior Técnico, University of Lisbon, Lisbon, Portugal, where he completed his Ph.D. degree with distinction and honour in Electrical and Computer Engineering in 2019. He is currently a Postdoctoral Fellow at the Department of Electrical and Computer Engineering, Queen’s University, where he works on different research projects funded by the Natural Sciences and Engineering Research Council of Canada (NSERC) and Mitacs, as well as private sector. His main research interests are machine learning and deep learning for biometrics, forensics, affective computing, and computer vision. He has contributed more than 35 papers in notable conferences and journals in his area and has been a reviewer for multiple top-tier conferences and journals in the field.  
\end{IEEEbiography}

\begin{IEEEbiography}[
{
\includegraphics[width=1in,height=1.25in,clip,keepaspectratio]{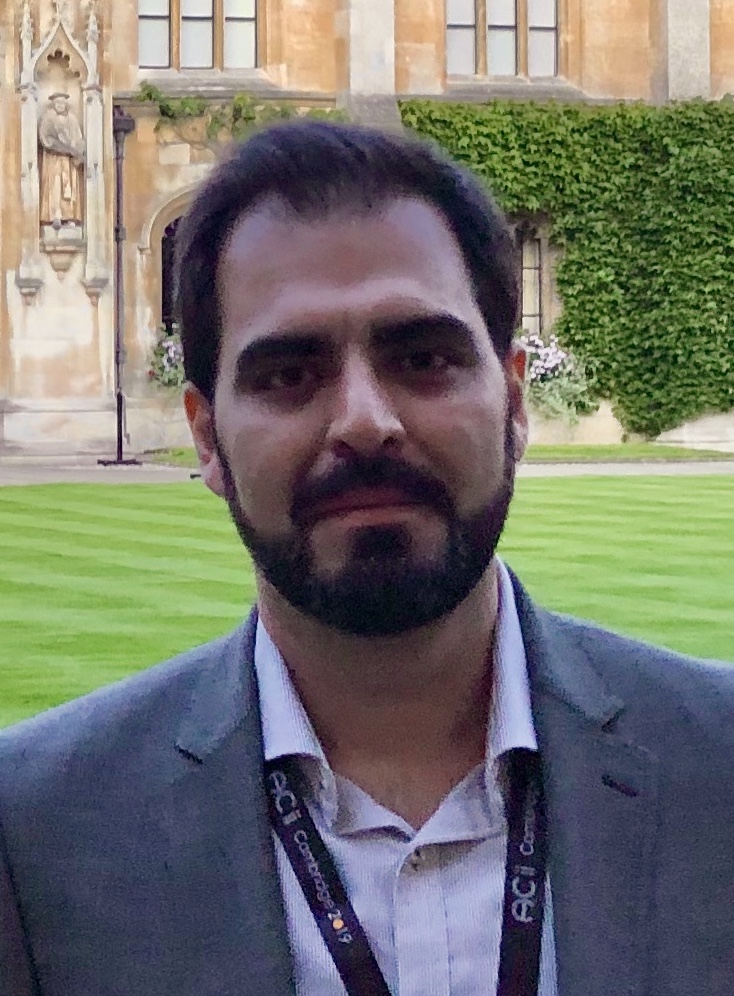}
}
]
{Ali Etemad} received the M.A.Sc. and Ph.D. degrees in Electrical and Computer Engineering from Carleton University, Ottawa, Canada, in 2009 and 2014, respectively. He is currently an Assistant Professor at the Department of Electrical and Computer Engineering, Queen’s University, Canada, where he leads the Ambient Intelligence and Interactive Machines (Aiim) lab. He is also a faculty member at Ingenuity Labs Research Institute. His main area of research is machine learning and deep learning focused on human-centered applications with wearables, smart devices, and smart environments. Prior to joining Queen’s, he held several industrial positions as lead scientist and director. He has co-authored over 90 articles and patents published/granted or under review, and has delivered over 15 invited talks regarding his work. He has been a member of program committees for several conferences in the field. He has been the recipient of a number of awards and grants. Dr. Etemad’s lab and research program have been funded by the Natural Sciences and Engineering Research Council (NSERC) of Canada, Ontario Centers of Excellence (OCE), Canadian Foundation for Innovation (CFI), Mitacs, and other organizations, as well as the private sector.

\end{IEEEbiography}

\end{document}